%% file: main.tex
\documentclass{article}

\usepackage{microtype}
\usepackage{graphicx}
\usepackage{subcaption}
\usepackage{booktabs} 

\usepackage{multirow}
\usepackage{makecell}

\usepackage{hyperref}

\usepackage[preprint]{icml2026}

\usepackage{xcolor}
\usepackage{tabularx}
\usepackage{pifont}
\usepackage[most]{tcolorbox}
\usepackage{amsmath}
\usepackage{amssymb}
\usepackage{mathtools}
\usepackage{amsthm}

\usepackage[capitalize,noabbrev]{cleveref}

\theoremstyle{plain}

\theoremstyle{definition}

\theoremstyle{remark}

\newcolumntype{L}{>{\raggedright\arraybackslash}X}
\usepackage[textsize=tiny]{todonotes}

\icmltitlerunning{Small Model as Master Orchestrator: Learning Unified Agent-Tool Orchestration with Parallel Subtask Decomposition}

\begin{document}

\twocolumn[
  \icmltitle{Small Model as Master Orchestrator: Learning Unified Agent-Tool \\ Orchestration with Parallel Subtask Decomposition}



  \icmlsetsymbol{equal}{*}

  \begin{icmlauthorlist}
    \icmlauthor{Wenzhen Yuan}{sjtu,equal}
    \icmlauthor{Wutao Xiong}{sichuan,equal}
    \icmlauthor{Fanchen Yu}{ailab}
    \icmlauthor{Shengji Tang}{cuhk}
    \icmlauthor{Ting Liu}{sjtu}
    \icmlauthor{Tao Chen}{fdu}
    \icmlauthor{Peng Ye}{cuhk,ailab}
    \icmlauthor{Yuzhuo Fu}{sjtu}
    \icmlauthor{Wanli Ouyang}{cuhk,ailab}
    \icmlauthor{Lei Bai}{ailab}
  \end{icmlauthorlist}

    \icmlaffiliation{sjtu}{Shanghai Jiao Tong University}
    \icmlaffiliation{fdu}{Fudan University}
    \icmlaffiliation{ailab}{Shanghai AI Lab}
    \icmlaffiliation{sichuan}{Sichuan University}
    \icmlaffiliation{cuhk}{The Chinese University of Hong Kong}


    \icmlcorrespondingauthor{Yuzhuo Fu}{yzfu@sjtu.edu.cn}
    \icmlcorrespondingauthor{Peng Ye}{yepeng@pjlab.org.cn}

  \icmlkeywords{Machine Learning, ICML}

  \vskip 0.3in
]



\printAffiliationsAndNotice{\icmlEqualContribution}

\begin{abstract}

Multi-agent systems (MAS) demonstrate clear advantages in tackling complex problems by coordinating diverse agents and external tools. However, most existing orchestration methods rely on static workflows or serial agent scheduling, and are further constrained by heterogeneous interface protocols between tools and agents.
This leads to high system complexity and poor extensibility.
To mitigate these issues, we propose Agent-as-Tool, a unified parallel orchestration paradigm that abstracts both agents and tools into a standardized, learnable action space with protocol normalization and explicit state feedback. Building on this paradigm, we train a lightweight orchestrator, ParaManager, which decouples planning decisions from subtask solving, enabling state-aware parallel subtask decomposition, delegation, and asynchronous execution.
For training, we adopt a two-stage ParaManager training pipeline. It improves robustness by incorporating supervised fine-tuning (SFT) trajectories equipped with recovery mechanisms, and further applies reinforcement learning (RL) to achieve an optimal balance among task success, protocol compliance, diversity, and reasoning efficiency. Experiments show that ParaManager achieves strong performance across multiple benchmarks and exhibits robust generalization under unseen model pools.
\end{abstract}

\input{sections/introduction}

\input{sections/related_work}

\input{sections/method}

\input{sections/experiment}

\input{sections/conclusion}




\bibliography{example_paper}
\bibliographystyle{icml2026}

\newpage
\appendix
\onecolumn


\input{sections/appendix_parameter}

\input{sections/appendix_reward}
\input{sections/appendix_case}
\clearpage
\input{sections/appendix_tools}
\clearpage
\input{sections/appendix_prompt}

\end{document}

%% file: sections/introduction.tex
\section{Introduction}

Large language models have made strong progress in natural language understanding, code generation, and complex reasoning~\cite{plaat2025reasoning, jiang2024survey}. Yet, for tasks that require long-horizon planning and substantial external knowledge, single-model end-to-end answering still suffers from systemic limitations~\cite{omar2025orchestrated, lin2025hallucinations}.
Consequently, MAS with tool-augmented reasoning have emerged as key directions. These frameworks leverage role specialization and external tools to provide a more reliable approach for complex tasks~\cite{hong2024metagpt, ye2025xmasbuildingmultiagentsystems}.

Many MAS rely on static orchestration, where fixed topologies or workflow templates are predefined and remain unchanged throughout execution~\cite{leong2025amas, yu2025dyntaskmas}. This design assumes that decomposition of subtasks and scheduling can be determined upfront, rather than being driven by intermediate evidence and observed failures~\cite{zhang2025aflowautomatingagenticworkflow, li2024autoflowautomatedworkflowgeneration}. 
In practice, orchestration is inherently stateful. The system must react to task states as well as execution signals~\cite{wang2025agentops}.
Without explicit state feedback mechanisms, faults at execution can propagate across intermediate steps, leading to brittle long-horizon trajectories~\cite{ToolOrchestra}.

To mitigate template rigidity, recent work explores learned routing and dynamic scheduling~\cite{dang2025multi, ToolOrchestra}. However, orchestration remains challenging because the action space is often constrained. On the one hand, workflows remain serial, limiting throughput and preventing multi-path execution~\cite{yu2025dyntaskmas, wang2025agentops}. On the other hand, systems depend on heterogeneous tools and predefined agent subtasks, which require a set of specialized agents for every potential function. This forces the orchestrator to handle incompatible tool schemas and failure semantics, hindering flexible repurposing of agents as the problem context evolves~\cite{zhang2025swarmagenticfullyautomatedagentic}.

We propose a unified parallel orchestration paradigm for agents and tools, \textbf{Agent-as-Tool}, which treats each agent as a tool with a standardized input-output protocol and an explicit execution state. This design places multi-agent collaboration and external tool use into a single learnable tool space, enabling orchestration through a unified action representation. Under this paradigm, we train a lightweight parallel orchestrator, the \textbf{ParaManager}, as the decision core of the system. 
At each round, ParaManager conditions on the current reasoning states to generate state-grounded subtasks, delegate them to appropriate agents and tools, schedule parallel executions, and determine whether to terminate.
This separation clarifies responsibilities: the ParaManager focuses on planning and decomposition, while invoked tools handle subproblem reasoning and evidence generation in isolated executions. This isolation reduces long-context interference and limits error propagation~\cite{ToolOrchestra,shen2026batchqueryprocessingoptimization}. 
The multi-round parallel orchestration provides redundancy across independent reasoning paths and reduces single-path failure risk compared with purely serial reasoning~\cite{wang2025surveyagentopscategorizationchallenges}. 

To realize this paradigm, we propose a two-stage framework for training a lightweight LLM as an effective orchestrator. We start with SFT on high-quality orchestration trajectories, and augment the training data with recoverable failure scenarios to teach ParaManager to repair format drift and handle execution faults. We further apply masked supervision on the ParaManager’s outputs to avoid overfitting to environment-side traces. Subsequently, we apply RL to align the model with system-level objectives. With a reward design that balances task success, protocol compliance, scheduling diversity, and efficiency, the framework yields a compact orchestrator capable of reliably coordinating agents and tools. 
In summary, our contributions are as follows:

\begin{figure*}[tbp]
    \centering
    \includegraphics[width=0.85\textwidth]{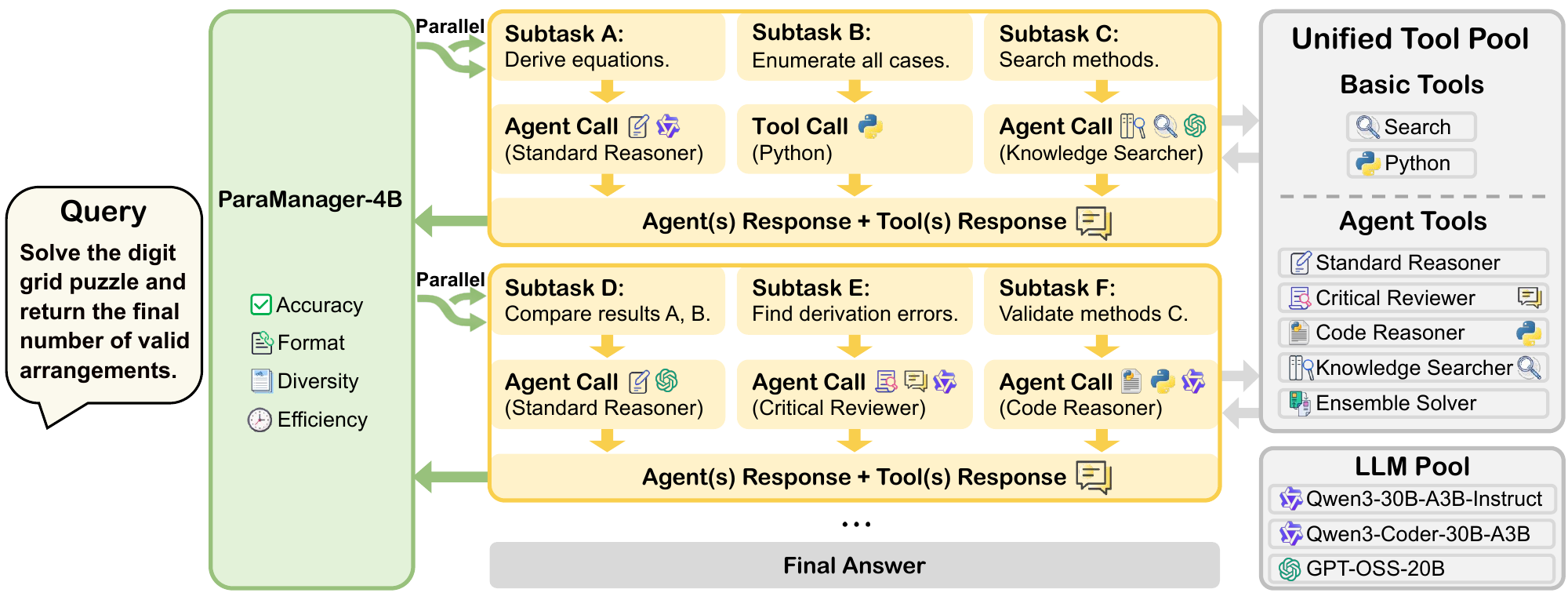}
    \caption{Multi-round, state-driven parallel orchestration: ParaManager iteratively decomposes the query into subtasks, launches parallel agent/tool calls via a unified tool pool in each round, and cross-checks and consolidates intermediate results to produce the final answer.}
    \vspace{-5pt}
    \label{fig:framework_our}
\end{figure*}

\textbf{Unified Agent-as-Tool paradigm with a lightweight ParaManager.} We unify agents and external tools into a single tool space via a standardized action schema and explicit feedback. By using a compact ParaManager to decouple high-level planning from low-level solving, we significantly reduce orchestration complexity.

\textbf{State-driven decomposition, subtask delegation, and parallel orchestration.} Our framework dynamically decomposes complex queries and delegates subtasks to appropriate agents and tools based on task characteristics. By enabling multi-round parallel tool calls, we improve reasoning accuracy through independent path validation and boost execution efficiency compared to purely serial workflows.

\textbf{ParaManager training with failure-recovery supervision and system-level RL objectives.} We develop a comprehensive training pipeline with SFT and RL that combines recoverable failure trajectories and masked supervision, and optimizes for task success, protocol compliance, diversity, and efficiency. This approach ensures the manager remains robust against execution instability and prevents collapse.

\textbf{Superior performance and cross-model generalization.} Extensive experiments on challenging benchmarks demonstrate that ParaManager consistently outperforms static workflows and strong orchestration baselines. Crucially, our framework exhibits robust generalization under model-pool shift, maintaining high efficacy when orchestrating unseen, larger-scale agent models without additional tuning.

%% file: sections/related_work.tex
\section{Related Work}
\label{sec:related_work}

\subsection{Automated Workflow Search and Optimization} Optimizing agent collaboration structures and workflow topology is a critical challenge in MAS \cite{yang2025topological, chen2025multi}. Since manual design is labor-intensive and limited in search scope \cite{hu2024automated}, automated optimization has become a research focus. Several works optimize workflows as code or graphs. AutoFlow \cite{li2024autoflowautomatedworkflowgeneration} leverages LLMs to iteratively construct task-specific workflows. AFlow \cite{zhang2025aflowautomatingagenticworkflow} formulates workflow search as a Monte Carlo Tree Search problem, exploring code-space operators to surpass manual designs. Evolutionary algorithms are also prevalent: EvoFlow \cite{zhang2025evoflowevolvingdiverseagentic} uses multi-objective evolution to generate diverse, heterogeneous workflows, balancing performance and cost. 
Finally, X-MAS \cite{ye2025xmasbuildingmultiagentsystems} targets heterogeneous node configuration, assigning specific models to optimal roles to enhance collective intelligence without altering topology.

\subsection{Agent Scheduling and Orchestration} 
Early systems often depended on static topologies or manual heuristics, limiting adaptability in dynamic environments \cite{jiang2025dynamic,leong2025amas,du-etal-2025-multi}. Research has shifted toward learning-based dynamic orchestration. To overcome static limitations, Puppeteer \cite{dang2025multi} models collaboration as a Markov Decision Process and trains an RL-based orchestrator to dynamically activate agents based on real-time states. ToolOrchestra \cite{ToolOrchestra} introduces an efficient RL-based method by training a lightweight orchestrator to coordinate diverse tools using outcome, efficiency, and preference rewards, reducing overhead while enhancing performance. 

Existing studies improve MAS either by optimizing workflow topology or by learning orchestration policies, yet most rely on heterogeneous interfaces and do not unify agent and tool calls within a single action space, making state-driven decomposition and parallel scheduling hard to learn end-to-end. As shown in Figure~\ref{fig:framework_our}, we address this gap with Agent-as-Tool, featuring tool-space unification, state-aware task decomposition, and parallel orchestration.

%% file: sections/method.tex
\section{Method}
\label{sec:method}


\subsection{Unified Agent-as-Tool parallel orchestration}
\label{sec:arch}

\paragraph{Tool-space unification.}
Let the tool library be $\mathcal{T}_{\text{tool}}=\{T_1,\dots,T_N\}$, where each $T_k$ can be an external tool such as search or code execution, or a sub-agent with its own independent context. All tools follow the same input-output protocol: the input is a structured parameter object $p$, the output is a structured result $v$, and each call carries an execution status $\sigma$. A single tool call returns a per-call observation
\begin{equation}
\begin{aligned}
o &\triangleq T_k(p) = (v,\; \sigma), \\
\sigma &\in \{\texttt{OK},\texttt{PARSE\_ERR},\texttt{EXEC\_ERR},\texttt{TIMEOUT}\}.
\end{aligned}
\end{equation}


\paragraph{ParaManager as a policy over actions.}
Given the original problem $q$, the system solves it within at most $H$ interaction rounds. At round $t$, the ParaManager receives state $s_t$ and generates reasoning text $r_t$ and an action $a_t$:
\begin{equation}
s_t = (q,\; \mathcal{H}_{t},\; b_t),\qquad
\mathcal{H}_{t}=\{(r_i,a_i,o_i)\}_{i=1}^{t-1}.
\end{equation}
where $b_t$ is the remaining reasoning budget and $o_i$ is the round-level tool observation at round $i$. For a tool-use round, the action is defined as a set of parallel tool calls
\begin{equation}
\label{eq:action}
\begin{gathered}
a_t = \{(k_{t,j}, p_{t,j})\}_{j=1}^{n_t}, \\
1 \le n_t \le n_{\max}, \quad k_{t,j} \in \{1,\dots,N\},
\end{gathered}
\end{equation}

where $k_{t,j}$ is the tool index for the $j$-th parallel call at round $t$, and $p_{t,j}$ is the corresponding parameter object for that call. In addition, the ParaManager may output a termination action $a_t=\texttt{final\_answer}$ to end the interaction.

\paragraph{Executor with structured feedback.}
To improve runtime stability, the executor validates each tool call independently and returns structured feedback. For the round-$t$ action $a_t=\{(k_{t,j},p_{t,j})\}_{j=1}^{n_t}$, we define a per-call validity indicator
\begin{equation}
\nu_{t,j}\triangleq \mathrm{Valid}(k_{t,j},p_{t,j})\in\{0,1\}.
\end{equation}
The executor executes only calls with $\nu_{t,j}=1$; otherwise, it skips the call and returns a parse error. Accordingly, the round-$t$ observation is a set of per-call observations
\begin{equation}
o_t=\{o_{t,j}\}_{j=1}^{n_t},\qquad o_{t,j}=(v_{t,j},\sigma_{t,j}).
\end{equation}
\begin{equation}
o_{t,j}=
\begin{cases}
T_{k_{t,j}}(p_{t,j}), & \nu_{t,j}=1,\\
(\varnothing,\texttt{PARSE\_ERR}), & \nu_{t,j}=0,
\end{cases}
\end{equation}
This design turns non-executable outputs into explicit state signals, allowing the ParaManager to correct formats, switch tools, or adjust parameters in later rounds, forming a recoverable closed-loop reasoning process.

\paragraph{Termination and final answer synthesis.}
The final answer can be generated by ParaManager itself or by a frozen summarizer $M_s$. In this work, we employ $M_s$ for trajectory-to-answer synthesis. When the ParaManager triggers termination, the system uses a summarization model $M_s$ to produce the final answer based on the history and the termination instruction $c_{\text{end}}$:
\begin{equation}
\hat{z}\sim M_s(\cdot\mid q,\mathcal{H}_T,r_T,c_{\text{end}}),
\end{equation}
where $\mathcal{H}_T=\{(r_i,a_i,o_i)\}_{i=1}^{T-1}$ is the trajectory history before round $T$, and $r_T$ is the reasoning text at round $T$.

With parallel tool calls and dynamic subtask assignment, the ParaManager can decompose the task into multiple subgoals, solve them in parallel, and enable redundancy and cross-checking across independent reasoning paths to reduce single-path failures (see Appendix~\ref{app:case_study} for case studies). 

\subsection{Supervised ParaManager Policy Initialization}
\label{sec:sft}

\begin{figure*}[tp]
    \centering
    \includegraphics[width=0.85\textwidth]{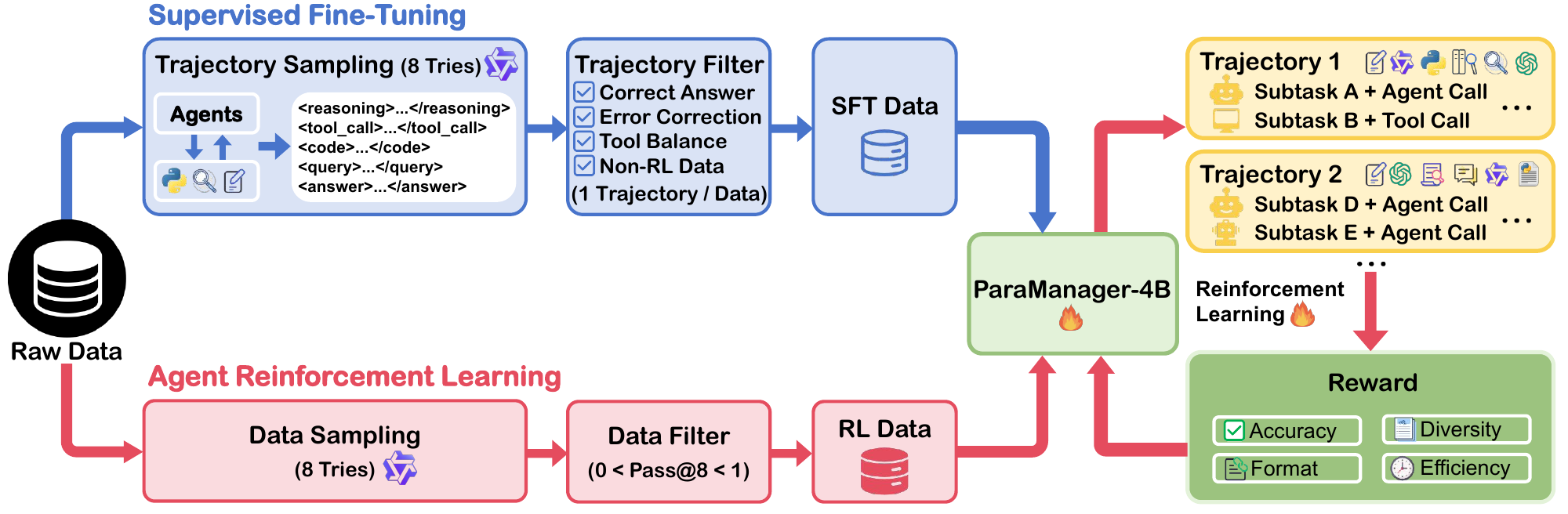}
    \caption{Two-stage training pipeline for ParaManager. \textbf{SFT:} sample multiple trajectories per instance, filter by correct final answer, error correction, and tool-balance, and retain one high-quality trajectory as supervision. \textbf{RL:} sample multiple rollouts, keep instances with mixed outcomes to ensure informative signals, and optimize ParaManager with rewards on accuracy, format, diversity, and efficiency.}
    \vspace{-5pt}
    \label{fig:framework_train}
\end{figure*}

Figure~\ref{fig:framework_train} overviews our two-stage training pipeline.

\paragraph{Trajectory dataset.}
We perform supervised learning on full orchestration trajectories generated online by the reasoning framework, denoted as $\mathcal{D}_{\text{sft}}=\{\tau\}$. Each trajectory records the interaction from problem input to termination:
\begin{equation}
\tau=\big(q,\; \{(r_i,a_i,o_i)\}_{i=1}^{T}\big),
\end{equation}

\paragraph{Recoverable error trajectories.}
To improve robustness to execution noise and formatting drift, we retain a fraction of recoverable error trajectories in $\mathcal{D}_{\text{sft}}$. Such trajectories include statuses such as \texttt{EXEC\_ERR}, and \texttt{TIMEOUT}, but the orchestrator recovers in later rounds by fixing call formats, switching tools, or adjusting parameters, and eventually terminates successfully. This training setup encourages the ParaManager to exploit structured failure signals for self-repair rather than collapsing after the first failure.

\paragraph{Masked SFT objective.}
The ParaManager generates reasoning text and action sequences autoregressively. Executor feedback $o_t$ is provided only as conditional context for the next round and is not used as a supervision target. We apply a token-level mask $m_\ell\in\{0,1\}$ to the linearized trajectory sequence and compute negative log-likelihood only on tokens produced by the ParaManager, while setting all environment-side tokens to zero:
\begin{equation}
\mathcal{L}_{\text{SFT}}(\theta)
= -\mathbb{E}_{\tau\sim \mathcal{D}_{\text{sft}}}
\sum_{\ell=1}^{|\mathrm{Seq}(\tau)|} m_\ell \log \pi_\theta\!\big(y_\ell \mid y_{<\ell}\big),
\end{equation}
where $\mathrm{Seq}(\tau)$ is the linearized training sequence of trajectory $\tau$, and $y_\ell$ is the $\ell$-th target token. The mask satisfies $m_\ell=1$ if and only if the token belongs to some round output $r_t$ or $a_t$; otherwise $m_\ell=0$. This objective focuses SFT on the generation rules of orchestration policies rather than fitting environment-side tool outputs.

\subsection{Reinforcement Learning for ParaManager Optimization}
\label{sec:rl}

We define the reward as a sum of four signals:
\begin{equation}
\label{eq:reward}
R(\tau)=R_{\text{task}}(\tau)+R_{\text{fmt}}(\tau)+R_{\text{div}}(\tau)+R_{\text{eff}}(\tau).
\end{equation}

\textbf{Task Accuracy reward.}
For tasks with verifiable answers, we directly compare the final output with the labeled answer. Let $\hat{z}$ be the final answer at termination, then
\begin{equation}
R_{\text{task}}(\tau)=\mathbb{I}\big[\hat{z}=\mathrm{GT}(q)\big].
\end{equation}

\textbf{Format reward.}
We enforce executability by requiring the ParaManager to follow the output template
\texttt{<reasoning>...</reasoning>} and \texttt{<tool\_call>...</tool\_call>}
in each round. Let $\mathrm{Fmt}(\cdot)\in\{0,1\}$ be the format checker for a single round output. The trajectory-level format reward is the per-round average:
\begin{equation}
R_{\text{fmt}}(\tau)=\frac{1}{T}\sum_{t=1}^{T}\mathrm{Fmt}(y_t),
\end{equation}
where $y_t$ is the full output of the ParaManager at round $t$.

\textbf{Diversity reward.}
We encourage parallel orchestration and exploration of tool combinations by defining two binary subrewards and averaging them:
\begin{equation}
R_{\text{div}}(\tau)=\frac{1}{2}\Big(R_{\text{par}}(\tau)+R_{\text{tool}}(\tau)\Big).
\end{equation}
The parallelism reward uses the per-round call count $n_t$ with a threshold $\theta_{\text{par}}$:
\begin{equation}
R_{\text{par}}(\tau)=\mathbb{I}\Big[\frac{1}{T}\sum_{t=1}^{T}n_t \ge \theta_{\text{par}}\Big].
\end{equation}
Let the effective tool set be $\mathcal{U}(\tau)$, which is defined as the union of unique tools excluding $\texttt{final\_answer}$
then with a threshold $\theta_{\text{tool}}$:
\begin{equation}
R_{\text{tool}}(\tau)=\mathbb{I}\big[|\mathcal{U}(\tau)|\ge \theta_{\text{tool}}\big].
\end{equation}

\textbf{Efficiency reward.}
We additionally regularize the policy toward concise reasoning and economical tool usage.
Inspired by the efficiency-aware formulations in DAPO~\cite{yu2025dapo} and Puppeteer, ~\cite{dang2025multi}, we impose soft budgets on both the total token length and the cumulative tool cost of a trajectory.
Let $L(\tau)$ be the total token length of the trajectory, and $C(\tau)$ be the total tool cost units along trajectory $\tau$.
We define target thresholds $L_{\text{tar}}$ for length and $C_{\text{tar}}$ for cost, with maximum thresholds set to twice the target values: $L_{\max} = 2L_{\text{tar}}$ and $C_{\max} = 2C_{\text{tar}}$.

The length reward component is given by
\begin{equation}
R_{\text{len}}(\tau)=
\begin{cases}
1, & L(\tau)\le L_{\text{tar}},\\
\max\Big(0,\; \frac{2L_{\text{tar}}-L(\tau)}{L_{\text{tar}}}\Big), & L(\tau)>L_{\text{tar}}.
\end{cases}
\end{equation}

The tool-cost component is defined analogously:
\begin{equation}
R_{\text{cost}}(\tau)=
\begin{cases}
1, & C(\tau)\le C_{\text{tar}},\\
\max\Big(0,\; \frac{2C_{\text{tar}}-C(\tau)}{C_{\text{tar}}}\Big), & C(\tau)>C_{\text{tar}}.
\end{cases}
\end{equation}

The overall efficiency reward averages the two:
\begin{equation}
R_{\text{eff}}(\tau)=\frac{1}{2}\Big(R_{\text{len}}(\tau)+R_{\text{cost}}(\tau)\Big).
\end{equation}

\paragraph{ParaManager policy optimization.}
We optimize the ParaManager policy $\pi_\theta$ with Group Relative Policy Optimization (GRPO)~\cite{shao2024deepseekmath} under tool-interleaved rollouts.
For each problem instance $q \sim P(Q)$, we sample a group of $G$ trajectories
$\mathcal{G}(q)=\{\tau^{(1)},\dots,\tau^{(G)}\}$ using the behavior policy $\pi_{\theta_{\text{old}}}$.
Each trajectory $\tau$ receives a scalar outcome reward $R(\tau)$ defined in Eq.~\ref{eq:reward}.
GRPO computes a group-relative normalized advantage
\begin{equation}
\label{eq:grpo_adv}
A(\tau)=
\frac{R(\tau)-\operatorname{mean}_{\tau'\in\mathcal{G}(q)}R(\tau')}
{\operatorname{std}_{\tau'\in\mathcal{G}(q)}R(\tau')+\delta},
\end{equation}
where $\delta$ is a small constant for numerical stability.

Different from standard GRPO where all tokens are generated by the policy, our trajectories interleave ParaManager-generated tokens with environment-side tokens emitted by non-policy tool outputs.
We introduce a token-level mask $m_\ell\in\{0,1\}$ (defined on the linearized sequence $\mathrm{Seq}(\tau)$) to exclude all non-policy tokens from optimization.
Concretely, we compute a masked, sequence-level importance ratio over ParaManager-generated tokens only
\begin{equation}
\label{eq:masked_ratio}
r_\theta(\tau)=
\prod_{\ell=1}^{|\mathrm{Seq}(\tau)|}
\left(
\frac{\pi_\theta(y_\ell\mid y_{<\ell})}
{\pi_{\theta_{\text{old}}}(y_\ell\mid y_{<\ell})}
\right)^{m_\ell},
\end{equation}
so that gradients are applied only to ParaManager outputs, while tool-generated tokens are treated purely as conditioning context.

The ParaManager is updated by maximizing a clipped surrogate objective
\begin{equation}
\label{eq:grpo_objective}
\max_{\theta}\;
\mathbb{E}_{q \sim P(Q)}
\left[
\frac{1}{G} \sum_{\tau \in \mathcal{G}(q)} \mathcal{L}_{\text{CLIP}}(\tau)
\right],
\end{equation}
where $\mathcal{L}_{\text{CLIP}}(\tau)$ is:
\begin{equation}
\label{eq:l_clip}
\min\Big(
r_\theta(\tau)A(\tau),\;
\operatorname{clip}\!\big(r_\theta(\tau), 1-\epsilon, 1+\epsilon\big)\,A(\tau)
\Big),
\end{equation}
and $\epsilon$ is the clipping hyperparameter.
Following DAPO~\cite{yu2025dapo}, we omit the KL penalty term to allow broader exploration of tool-calling strategies.

%% file: sections/experiment.tex
\section{Experiment}
\label{sec:experiment}

\input{tables/main_table.tex}

\subsection{Tools}


\paragraph{Basic tools.}
We execute model-generated Python code via the sandbox interpreter~\cite{bytedanceseedfoundationcodeteam2025fullstackbenchevaluatingllms}. The executor returns stdout/stderr verbatim, so the model can iterate on failing programs rather than restarting from scratch. For retrieval, we follow Search-R1~\cite{jin2025search,jin2025empirical} with e5-base-v2~\cite{wang2022text} as the retriever and Wikipedia as the corpus.

\paragraph{Agent tools.}
Agent tools take a subtask description and a model identifier and return a solution under a fixed I/O protocol:
\textbf{Standard Reasoner} produces direct answer without external tool calls.
\textbf{Critical Reviewer} inspects the reasoning history, detects potential issues, and suggests fixes.
\textbf{Code Reasoner} writes and runs Python code for a subtask.
\textbf{Knowledge Searcher} searches for evidence for a subtask.
\textbf{Ensemble Solver} performs multi-path sampling and returns an aggregated result.
\textbf{Final Answer} End the reasoning and return the final answer
Detailed definitions and tool cost units are provided in Appendix~\ref{app:tools}.

\subsection{Data Collection}

\paragraph{RL Data.}
We derive RL instances from DAPO-Math~\cite{yu2025dapo}. For each question, we sample eight outputs using Qwen3-30B-A3B-Instruct-2507~\cite{qwen3technicalreport} and discard questions whose eight samples are all correct or all incorrect: the former yields near-zero preference signal, while the latter may reflect unsolvable or low-quality questions. The resulting RL set contains 3,742 instances.

\paragraph{SFT Data.}
We collect SFT trajectories to initialize ParaManager. We run Qwen3-30B-A3B-Instruct-2507 on the DAPO-Math training split using our framework and keep trajectories that end with the correct answer. To avoid leakage, we remove any question that overlaps with the RL set. We sample eight trajectories per instance and keep at most one, prioritizing runs that recover from tool failures. To avoid tool imbalance, we enforce a simple distribution constraint so that no single tool dominates the retained set. This yields 4,470 SFT instances.

\subsection{Training Configuration}
ParaManager is initialized from Qwen3-4B-Instruct-2507. The agent pool includes Qwen3-30B-A3B-Instruct-2507, Qwen3-Coder-30B-A3B-Instruct~\cite{qwen3technicalreport}, and GPT-OSS-20B~\cite{agarwal2025gpt}. Training is implemented with Verl~\cite{sheng2024hybridflow}. We run SFT for 200 steps with batch size 32, followed by RL for 50 steps with batch size 16. Each instance is solved within at most 12 interaction rounds, and each round allows up to four parallel tool calls. During RL, we sample eight trajectories per instance and optimize ParaManager with GRPO. We set the diversity thresholds to $\theta_{\text{par}}=1.25$ and $\theta_{\text{tool}}=3$. For efficiency regularization, we use target budgets $L_{\text{tar}}=12{,}288$ tokens and $C_{\text{tar}}=8$ tool-cost units. We use GPT-OSS-20B as the termination model $M_s$ for trajectory-to-answer synthesis. Hyperparameter details are reported in Appendix~\ref{app:training_configuration}. All experiments are conducted on 12 NVIDIA H200 GPUs.

\subsection{Baselines and Evaluation}

\paragraph{Pure model baselines.} This group measures the native performance of a single model. We also include two test-time scaling strategies, majority voting at 8 samples and self-refinement~\cite{madaan2023self}. In addition, we evaluate tool-augmented single-model performance of each model.

\paragraph{Static workflow baselines.} This group uses predefined static workflows for task scheduling. Representative methods include Router~\cite{ong2024routellm}, Meta Agent Search~\cite{hu2024automated}, and EvoFlow~\cite{zhang2025evoflowevolvingdiverseagentic}.
For Router, we build a problem bank from the DAPO-Math training split, retrieve similar problems using k-nearest neighbors, and select the output from the model that scores best on the retrieved set.
For Meta Agent Search, we use Qwen3-30B-A3B-Instruct-2507 as the base model to construct the workflow.
For EvoFlow, we construct the workflow using the full model pool.

\paragraph{Serial orchestration baselines.} This group evaluates dynamic yet serial orchestration strategies.
We compare against Puppeteer~\cite{dang2025multi} and ToolOrchestra~\cite{ToolOrchestra}. We use the officially released weights with our tool suite to ensure a fair comparison.

\paragraph{Benchmarks.} 
We evaluate both in-domain math benchmarks and out-of-domain general benchmarks. For in-domain evaluation, we adopt four math datasets: AIME24, AIME25, AMC~\cite{li2024numinamath}, and HMMT25~\cite{balunovic_srimatharena_2025}, which contain high-difficulty problems. For out-of-domain evaluation, we further include GPQA~\cite{rein2024gpqa}, MMLU-pro~\cite{wang2024mmlu}, LiveCodeBenchv5 (LCBv5) and LiveCodeBenchv6 (LCBv6)~\cite{jain2024livecodebench} to measure cross-domain generalization on broad knowledge, reasoning, and coding-oriented tasks. For LiveCodeBench, we follow the official evaluation code and use the standard prompts for all models to ensure a fair comparison.

To ensure a fair comparison across methods, all baselines share the same tool suite and model pool, and we set the maximum generation length to 24,576 tokens and temperature of 1.0 for every model invocation. For each benchmark instance, we run 8 independent samples and report the mean accuracy (mean@8).

\begin{figure}[tbp]
    \centering
    \includegraphics[width=0.45\textwidth]{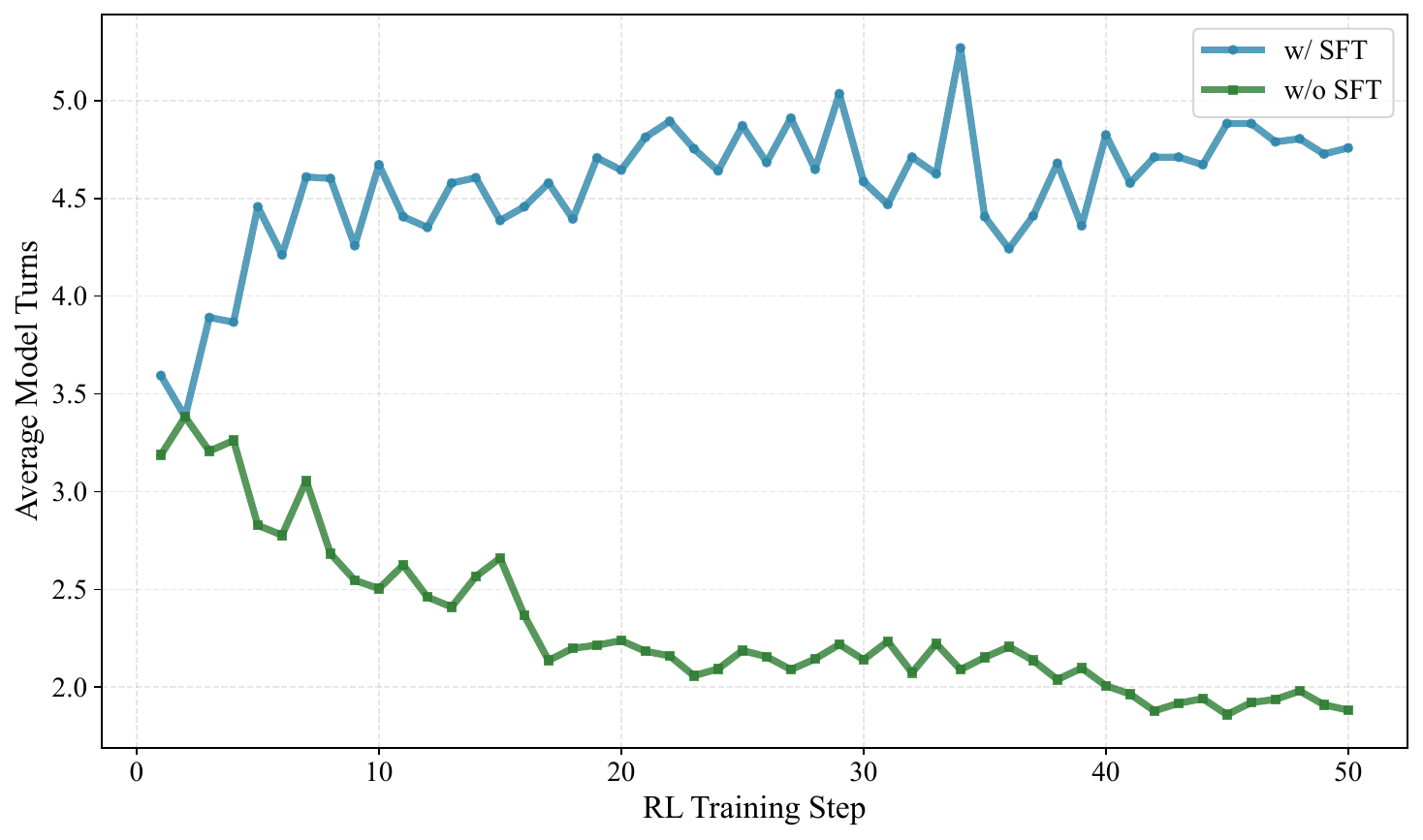}
    \caption{The curve of changes in the average number of model turns in the RL stage with/without SFT initialization}
    \label{fig:sft_necessary}
\end{figure}

\subsection{Experimental Results}
\label{subsec:experimental_results}
Table~\ref{tab:main_results} summarizes the main results across in-domain math and out-of-domain benchmarks.
ParaManager achieves the best overall average of 70.48, outperforming all backbone-only baselines and prior orchestration methods. Compared with the strongest backbone GPT-OSS-20B with Majority Vote of 66.88, ParaManager improves by 3.60 points. It also outperforms the best static-workflow baseline EvoFlow at 65.76 by 4.72 points and the serial orchestration baseline ToolOrchestra at 66.21.

The gains are consistent across domains. ParaManager ranks first on all four math benchmarks, reaching 86.67 on AIME24, 84.17 on AIME25, 95.18 on AMC, and 63.33 on HMMT25. On out-of-domain tasks, it scores 72.10 on GPQA and 81.43 on MMLU-pro. It remains competitive on code benchmarks. These findings suggest the learned scheduling policy transfers beyond math without over-specializing to a single task style.

Ablations highlight the role of training and orchestration design. ParaManager-SFT reaches 63.43 on average, while the full ParaManager reaches 70.48, showing a 7.05 point gain from RL. Restricting the pool to a single backbone of Qwen3-30B-A3B-Instruct-2507 in ParaManager-Mono reduces the average to 65.54, a drop of 4.94 points, supporting the value of heterogeneous agents. Enforcing serial execution in ParaManager-Serial further reduces performance to 66.48, 4.00 points below ParaManager. This pattern indicates that parallel calls improve reliability through cross-verification and stronger intermediate evidence.

\section{Analysis}
\label{sec:analysis}

\begin{figure}[tbp]
    \centering
    \includegraphics[width=0.45\textwidth]{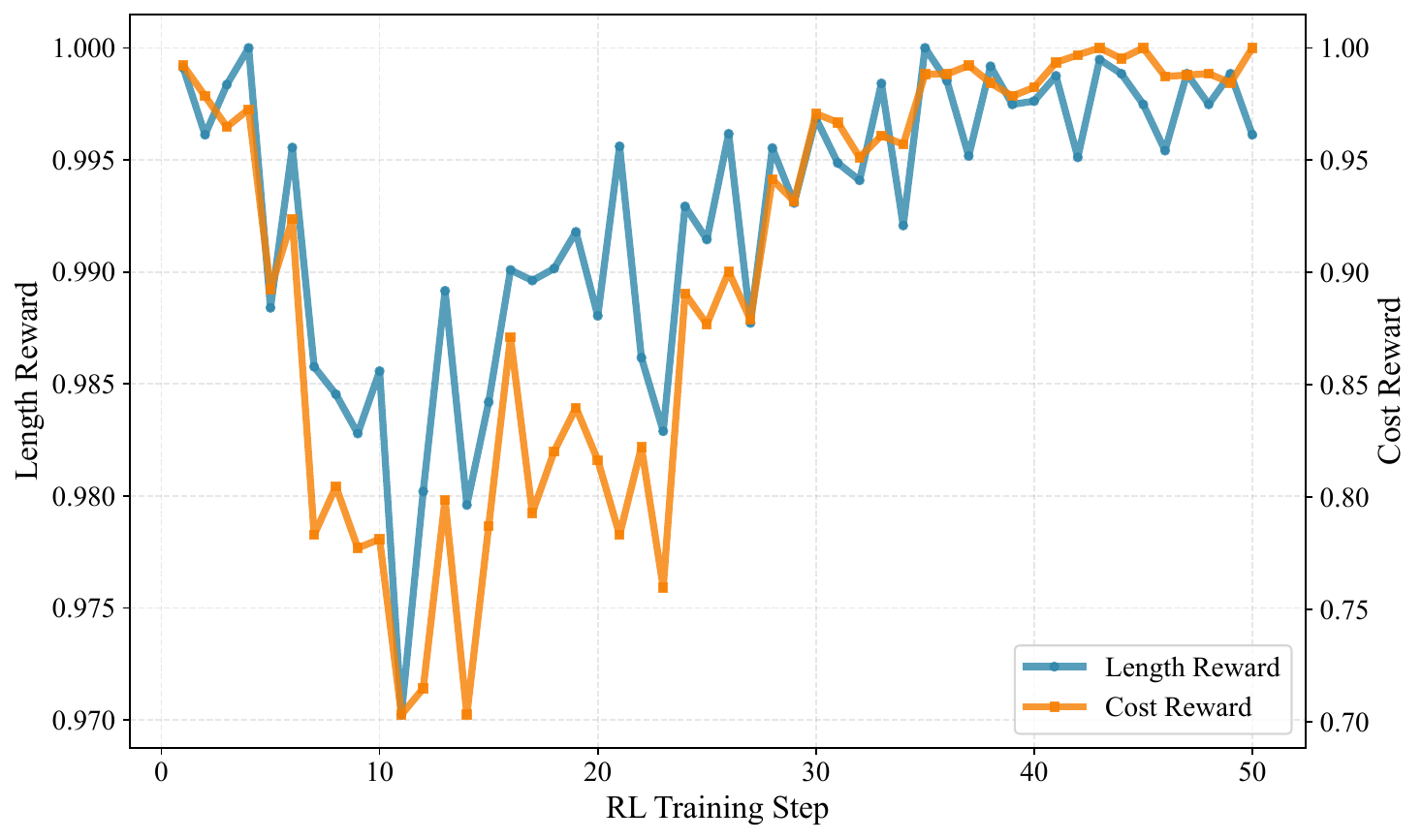}
    \caption{Tool cost reward and length reward curves during the training process.}
    \label{fig:cost_analysis}
\end{figure}

\subsection{Why SFT Initialization Is Necessary}


In the Agent-as-Tool paradigm, the orchestrator must generate structured calls that satisfy format and semantic constraints in order to be executable and to yield valid environment feedback. Starting RL directly from a base model is often unstable because early policies rarely produce executable actions consistently, causing most trajectories to terminate with parsing or execution failures and resulting in prolonged sparse rewards. Under such limited supervision, optimization can drift toward locally stable but degenerate behaviors, most notably avoiding tool usage and reverting to single-turn direct answering, which leads to the collapse shown in Figure~\ref{fig:sft_necessary}. SFT initialization allows RL to focus on system-level objectives rather than learning action schemas and failure semantics from scratch. Moreover, by including trajectories that recover from errors, the model learns to leverage structured failure signals for self-repair, reducing invalid calls and improving the stability of subsequent RL.

\subsection{Efficiency Reward Analysis}
\label{subsec:cost_analysis}

We analyze the role of the efficiency reward during training. Figure~\ref{fig:cost_analysis} shows the tool cost reward and the length reward over training. In the early stage, both signals drop rapidly, indicating that the policy explores more aggressively by invoking tools more often and producing longer reasoning traces. As training progresses, the cost and length signals gradually rise and then stabilize. This suggests that the model learns to reduce unnecessary tool invocations and redundant reasoning, controlling computational cost while maintaining output quality.

\begin{figure}[tbp]
    \centering
    \includegraphics[width=0.45\textwidth]{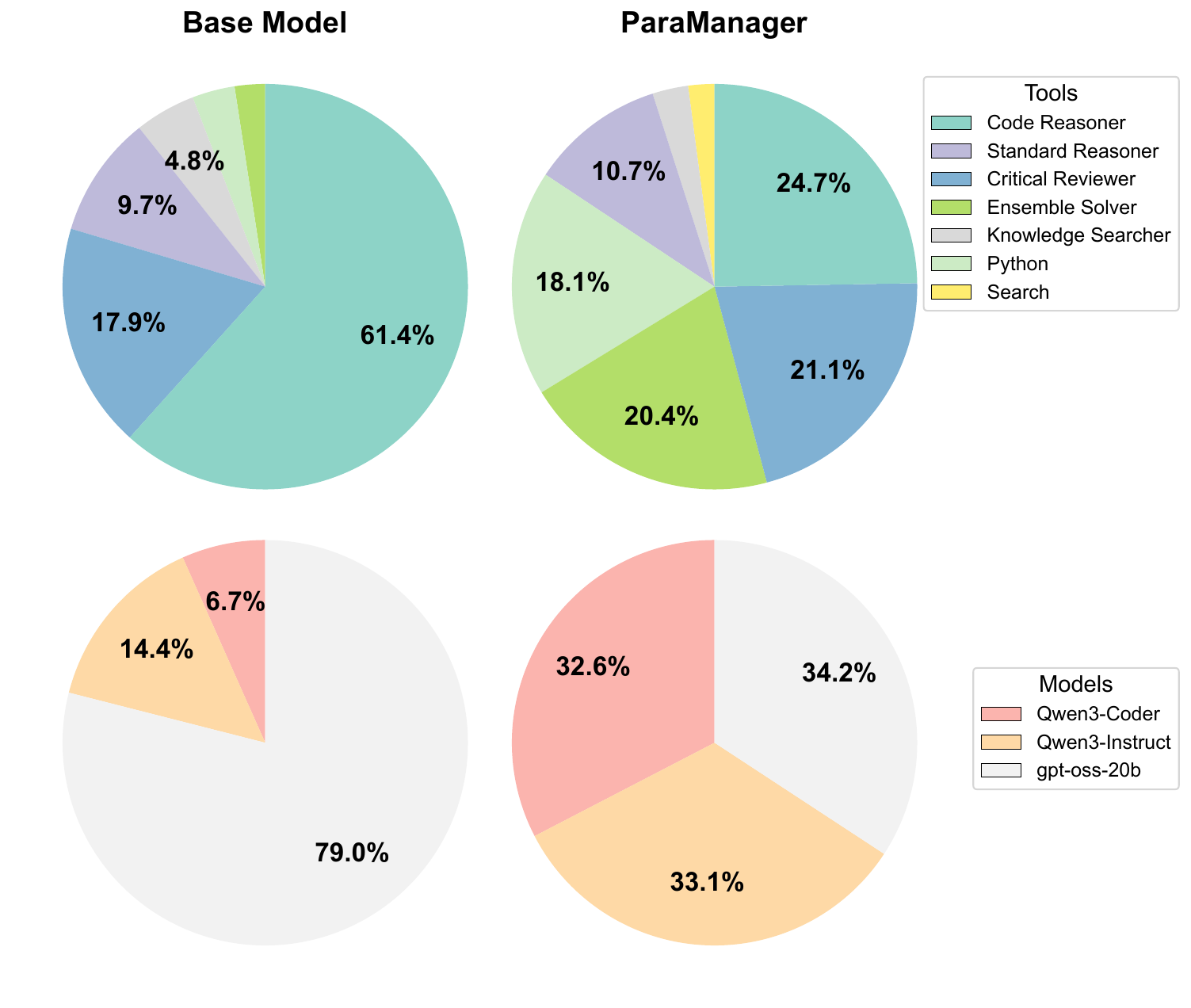}
    \caption{Model and tool distribution of the untrained base model Qwen3-4B-Instruct-2507 and the trained ParaManager. The top row shows tool distribution, and the bottom row shows model distribution.}
    \label{fig:model_tool_distribution}
\end{figure}

\subsection{Model and Tool Distribution}
\label{subsec:tool_use_analysis}

We study how training reshapes model routing and tool usage. Figure~\ref{fig:model_tool_distribution} compares the untrained base model with the trained ParaManager.
For model selection, the untrained base model routes most calls to GPT-OSS-20B, with Qwen3-Instruct and Qwen3-Coder rarely used. After training, ParaManager produces a near-uniform split across the three models of 34.2\%, 33.1\%, and 32.6\%, indicating more systematic load balancing via cross-model complementarity.

Tool usage shows a similar shift. In the untrained setting, \texttt{code\_reasoner} dominates. After training, usage becomes more diverse: \texttt{code\_reasoner} drops to 24.7\%, while \texttt{critical\_reviewer}, \texttt{ensemble\_solver}, and \texttt{python} rise to 21.1\%, 20.4\%, and 18.1\%. Training promotes tool composition driven by subtask requirements rather than reliance on a single default tool.

\subsection{Budget-Matched Efficiency Comparison}
\label{subsec:budget_matched_efficiency}

\begin{table}[tbp]
    \centering
    \caption{Average accuracy on math tasks and average token consumption with latency for different methods.}
    \label{tab:budget_matched}
    \resizebox{\columnwidth}{!}{
    \begin{tabular}{lccc}
        \toprule
        Method & ACC (\%) & Token & Latency (s) \\
        \midrule
        \multicolumn{4}{l}{\textbf{Orchestration Baselines}} \\
        Meta Agent Search & 73.96 & 65478 & 602.16 \\
        EvoFlow & 76.97 & 58168 & 680.38 \\
        Puppeteer & 74.22 & 45932 & 558.19 \\
        ToolOrchestra & 76.97 & 65815 & 756.88 \\
        \midrule
        \multicolumn{4}{l}{\textbf{Ours}} \\
        ParaManager-Serial & 76.51 & 46801 & 660.32 \\
        ParaManager & 82.34 & 55063 & 528.84 \\
        \bottomrule
    \end{tabular}
    }
\end{table}

We further investigate whether the observed performance gains stem from increased computational overhead. Table~\ref{tab:budget_matched} shows that ParaManager reaches the highest accuracy of 82.34\% with 55,063 tokens and the lowest latency of 528.84s. Importantly, these gains are not due to cost expansion.
Compared to our serial variant, ParaManager-Serial reduces token usage to 46,801 but suffers from lower accuracy at 76.51\% and increased latency of 660.32 seconds, confirming that our parallel scheduling improves both reliability and wall-clock efficiency.

\subsection{Generalize to Unseen Model Pool}
\label{subsec:generalization}

We evaluate generalization by fully replacing the model pool with three larger models that are unseen during training, while keeping the tool set and agent roles fixed. We refresh the model descriptions and let the manager orchestrate over this new pool. Table~\ref{tab:generalization} shows that ParaManager achieves the best average accuracy of 86.52, surpassing other orchestration baselines such as Puppeteer and ToolOrchestra. It also exceeds the best single-model average of 84.93. At the task level, ParaManager matches the top score on AIME24. The gains are strongest on AIME25, and ParaManager stays competitive with the strongest single-model voting results across benchmarks. These results indicate that the learned scheduling policy transfers across model pools and consistently benefits from model diversity.

\begin{table}[tbp]
    \centering
    \caption{Generalization evaluation on an unseen model pool across math benchmarks.}
    \label{tab:generalization}
    \resizebox{\columnwidth}{!}{
    \begin{tabular}{lccccc}
        \toprule
        Method & AIME24 & AIME25 & AMC & HMMT25 & AVG \\
        \midrule
        \multicolumn{6}{l}{\textbf{Gemini-2.5-flash}} \\
        Base & 78.75 & 67.08 & 92.47 & 60.83 & 74.78 \\
        Majority Vote & \textbf{86.67} & \textbf{80.00} & \textbf{96.39} & \textbf{76.67} & \textbf{84.93} \\
        Self Refine & \textbf{86.67} & 76.67 & 93.98 & 73.33 & 82.66 \\
        \midrule
        \multicolumn{6}{l}{\textbf{Qwen3-235B-A22B-Instruct-2507}} \\
        Base & 80.00 & 70.00 & 94.73 & 50.42 & 73.79 \\
        Majority Vote & \textbf{85.83} & \textbf{76.67} & \textbf{95.18} & 50.00 & \textbf{76.92} \\
        Self Refine & 83.33 & \textbf{76.67} & 94.58 & \textbf{50.83} & 76.35 \\
        \midrule
        \multicolumn{6}{l}{\textbf{GPT-OSS-120B}} \\
        Base & 76.25 & 78.75 & 91.27 & 48.33 & 73.65 \\
        Majority Vote & \textbf{83.33} & \textbf{91.67} & \textbf{95.18} & \textbf{66.67} & \textbf{84.21} \\
        Self Refine & \textbf{83.33} & 90.00 & 92.77 & 60.00 & 81.53 \\
        \midrule
        \multicolumn{6}{l}{\textbf{Model Orchestration}} \\
        Puppeteer & 80.00 & 76.67 & 91.87 & 73.33 & 80.47 \\
        ToolOrchestra & 83.33 & 78.75 & 92.62 & 72.92 & 81.90 \\
        \textbf{ParaManager} & \textbf{86.67} & \textbf{90.00} & \textbf{94.43} & \textbf{75.00} & \textbf{86.52} \\
        \bottomrule
    \end{tabular}
    }
\end{table}

%% file: tables/main_table.tex
\begin{table*}[t]
\centering
\caption{Accuracy (\%) under mean@8 on in-domain math benchmarks and out-of-domain benchmarks.
We report the average across all datasets in the last column. The best results are highlighted in \textbf{bold}, and the second-best results are \underline{underlined}.}
\label{tab:main_results}
\small
\setlength{\tabcolsep}{4pt}
\renewcommand{\arraystretch}{1.05}
\resizebox{\textwidth}{!}{
\begin{tabular}{llccccccccc}
\toprule
\multirow{2}{*}{} & \multirow{2}{*}{Method} &
\multicolumn{4}{c}{In-domain} &
\multicolumn{4}{c}{Out-of-domain} &
\multirow{2}{*}{AVG} \\
\cmidrule(lr){3-6}\cmidrule(lr){7-10}
& & {AIME24} & {AIME25} & {AMC} & {HMMT25} & {GPQA} & {MMLU-pro} & {LCBv5} & {LCBv6} & \\
\midrule

\multirow{4}{*}{\makecell[l]{Qwen3-4B\\Instruct-2507}}
& Base           & 68.33 & 51.67 & 86.90 & 34.17 & 59.09 & 74.82 & 36.60 & 39.95 & 56.44 \\
& Base+Tool      & 68.33 & 52.50 & 88.25 & 34.17 & 59.34 & 74.11 & 37.15 & 39.52 & 56.67 \\
& Majority Vote  & 83.33 & 56.67 & 91.57 & 40.00 & 62.63 & 72.86 & 35.83 & 39.54 & 60.30 \\
& Self Refine    & 80.00 & 60.00 & 89.61 & 37.50 & 60.10 & 75.71 & 38.36 & 40.12 & 60.17 \\

\addlinespace[2pt]

\multirow{4}{*}{\makecell[l]{Qwen3-30B\\A3B-Instruct-2507}}
& Base           & 77.50 & 63.33 & 91.87 & 42.50 & 63.13 & 76.96 & 39.12 & 42.10 & 62.06 \\
& Base+Tool      & 66.67 & 58.33 & 90.36 & 43.33 & 59.97 & 77.86 & \textbf{39.76} & 41.25 & 59.69 \\
& Majority Vote  & 80.00 & 66.67 & \textbf{95.18} & 43.33 & 64.65 & 78.57 & 39.28 & 42.45 & 63.77 \\
& Self Refine    & 79.17 & 67.50 & 92.17 & 43.75 & 64.39 & 76.79 & 39.42 & \textbf{43.25} & 63.30 \\

\addlinespace[2pt]

\multirow{4}{*}{GPT-OSS-20B}
& Base           & 73.33 & 70.42 & 90.66 & 47.50 & 62.12 & 79.64 & 33.53 & 37.43 & 61.83 \\
& Base+Tool      & 76.67 & 70.42 & 93.98 & 56.67 & 65.40 & 80.00 & 34.58 & 40.87 & 64.82 \\
& Majority Vote  & 83.33 & \underline{80.00} & 92.17 & 56.67 & \underline{71.21} & 80.00 & 34.50 & 37.20 & \underline{66.88} \\
& Self Refine    & 83.33 & 79.17 & 92.17 & 52.08 & 69.44 & 80.00 & 37.20 & 40.20 & 66.70 \\

\addlinespace[2pt]

\multirow{4}{*}{\makecell[l]{Qwen3-Coder\\30B-A3B-Instruct}}
& Base           & 28.33 & 21.67 & 58.73 & 10.83 & 53.47 & 77.14 & 34.76 & 39.68 & 40.58 \\
& Base+Tool      & 43.33 & 33.33 & 61.45 & 23.33 & 57.45 & 75.71 & 36.20 & \underline{42.80} & 46.70 \\
& Majority Vote  & 33.33 & 26.67 & 61.45 & 13.33 & 55.05 & 80.00 & 35.20 & 40.20 & 43.15 \\
& Self Refine    & 34.17 & 26.67 & 60.24 & 12.50 & 56.82 & 80.36 & 37.60 & 40.20 & 43.57 \\

\midrule

\multirow{3}{*}{\makecell[l]{Static Workflow}}
& Router
                 & 77.50 & 71.25 & 91.87 & 47.92 & 59.22 & 80.00 & 39.15 & 41.25 & 63.52 \\
& \makecell[l]{Meta Agent Search}
                 & \underline{84.17} & 72.92 & \underline{94.58} & 44.17 & 56.82 & 79.46 & 32.08 & 35.16 & 62.42 \\
& \makecell[l]{EvoFlow}
                 & \underline{84.17} & 79.17 & 92.47 & 52.08 & 68.18 & 79.82 & 33.06 & 37.12 & 65.76 \\

\addlinespace[2pt]
\midrule

\multirow{2}{*}{\makecell[l]{Serial Orchestration}}
& \makecell[l]{Puppeteer}
                 & 75.83 & 70.00 & 93.98 & 57.08 & 64.77 & \underline{80.54} & 39.11 & 41.06 & 65.30 \\
& \makecell[l]{ToolOrchestra}
                 & 80.00 & 79.17 & 90.36 & 58.33 & 66.41 & 79.46 & 37.64 & 38.27 & 66.21 \\

\midrule

\multirow{4}{*}{\makecell[l]{ParaManager-4B}}
& \makecell[l]{ParaManager-SFT}
                 & 80.00 & 66.67 & 92.02 & 56.67 & 66.29 & \textbf{81.43} & 30.80 & 33.58 & 63.43 \\
& \makecell[l]{ParaManager-Mono}
                 & 83.33 & 70.00 & 92.17 & 50.00 & 67.49 & 79.64 & 39.20 & 42.50 & 65.54 \\
& \makecell[l]{ParaManager-Serial}
                 & 77.50 & 76.67 & 91.87 & \underline{60.00} & 65.03 & 80.00 & 39.43 & 41.38 & 66.48 \\
& \makecell[l]{ParaManager}
                 & \textbf{86.67} & \textbf{84.17} & \textbf{95.18} & \textbf{63.33} & \textbf{72.10} & \textbf{81.43} & \underline{39.45} & 41.51 & \textbf{70.48} \\

\bottomrule
\end{tabular}
}

\end{table*}

%% file: sections/conclusion.tex
\section{Conclusion}
We proposed Agent-as-Tool and trained a lightweight parallel orchestrator, ParaManager, to unify agent and tool calls under a standardized protocol with explicit state feedback. Through SFT initialization and RL optimization, ParaManager learns state-aware parallel decomposition, delegation, and verification, improving robustness and efficiency. Experiments show consistent gains over strong single-model and orchestration baselines on both in-domain math and out-of-domain benchmarks, and the approach generalizes under model-pool replacement.

%% file: sections/appendix_parameter.tex
\section{Detailed Training Configurations}
\label{app:training_configuration}

Tables~\ref{tab:sft_hyperparameters} and~\ref{tab:rl_hyperparameters} summarize the key hyperparameters for the SFT and RL stages, respectively.

\begin{table}[ht]
\centering
\caption{Key Hyperparameters for SFT Phase}
\label{tab:sft_hyperparameters}
\small
\begin{tabular}{lc}
\toprule
\textbf{Hyperparameter} & \textbf{Value} \\
\midrule
\textit{Data \& Batch} & \\
train\_batch\_size & 32 \\
max\_length & 49152 \\
total\_training\_steps & 200 \\
\midrule
\textit{Optimizer \& LR} & \\
optimizer & AdamW \\
learning\_rate & 1e-5 \\
lr\_scheduler & cosine \\
\midrule
\textit{Multi-turn Tool Use} & \\
max\_assistant\_turns & 12 \\
max\_parallel\_calls & 4 \\

\bottomrule
\end{tabular}
\end{table}

\begin{table}[ht]
\centering
\caption{Key Hyperparameters for RL (GRPO) Phase}
\label{tab:rl_hyperparameters}
\small
\begin{tabular}{lc}
\toprule
\textbf{Hyperparameter} & \textbf{Value} \\
\midrule
\textit{Data \& Length} & \\
train\_batch\_size & 16 \\
max\_prompt\_length & 24576 \\
max\_response\_length & 24576 \\
\midrule
\textit{Optimizer \& Parallelism} & \\
actor\_learning\_rate & 2e-6 \\
tensor\_model\_parallel\_size & 4 \\
\midrule
\textit{GRPO Config} & \\
adv\_estimator & grpo \\
mini\_batch\_size & 8 \\
clip\_ratio (low / high) & 0.2 / 0.28 \\
rollout\_temperature & 1.0 \\
rollout\_top\_p & 1.0 \\
\midrule
\textit{Rollout \& Budget} & \\
rollout\_n (G) & 8 \\
total\_training\_steps & 50 \\
\midrule
\textit{Reward Thresholds \& Budgets} & \\
theta\_par ($\theta_{\text{par}}$) & 1.25 \\
theta\_tool ($\theta_{\text{tool}}$) & 3 \\
length\_target ($L_{\text{tar}}$) & 12288 \\
cost\_target ($C_{\text{tar}}$) & 8 \\
\midrule
\textit{Multi-turn Tool Use} & \\
max\_assistant\_turns & 12 \\
max\_parallel\_calls & 4 \\
max\_tool\_response\_length & 4096 \\
\bottomrule
\end{tabular}
\end{table}

%% file: sections/appendix_reward.tex
\section{Reward design}
\subsection{Diversity Reward Analysis}
\label{subsec:diversity_reward_analysis}



\begin{figure}[tbp]
    \centering
    \begin{subfigure}[b]{0.48\textwidth}
        \centering
        \includegraphics[width=\linewidth]{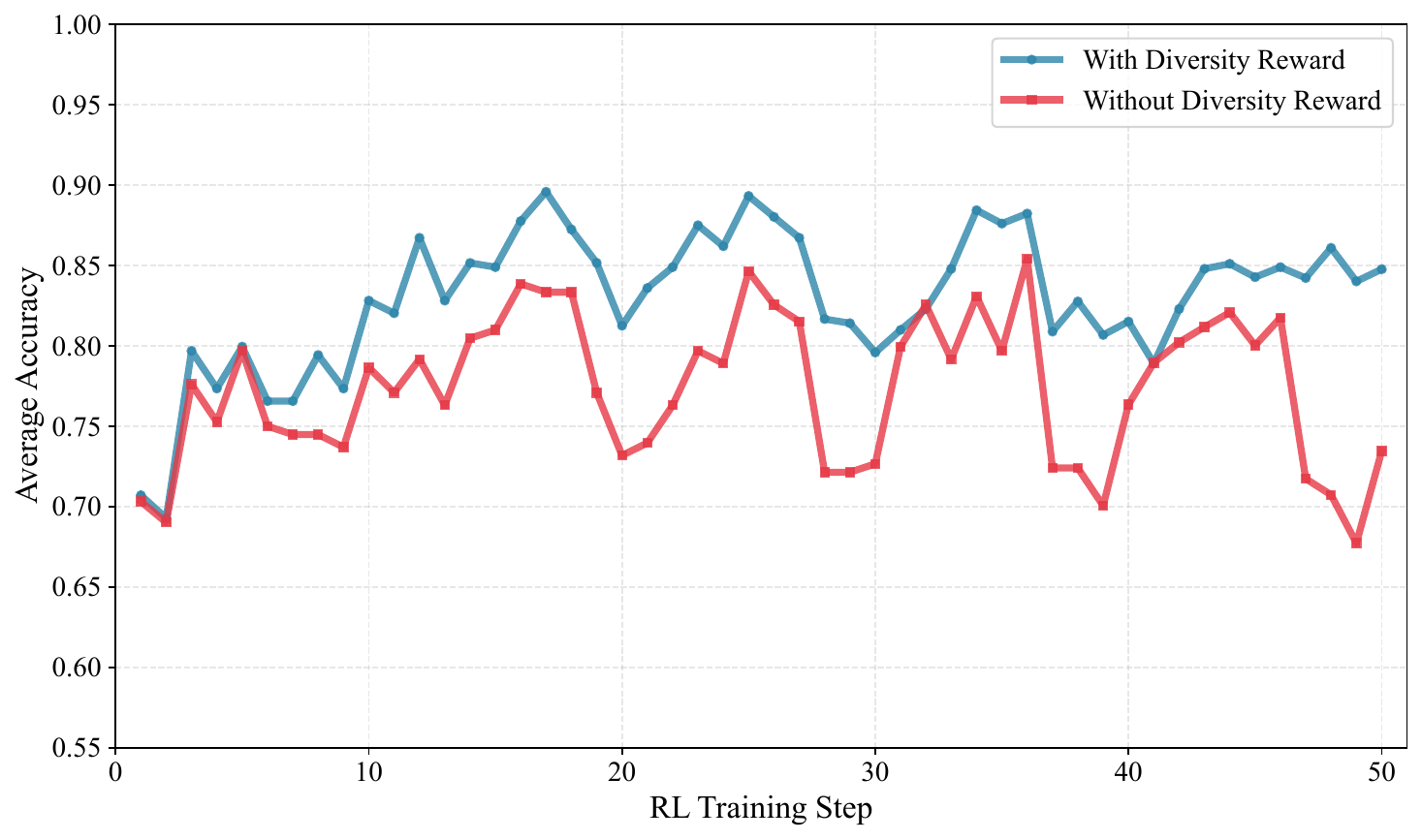}
        \caption{Average accuracy on training set.}
        \label{fig:diversity_reward_perf_curve}
    \end{subfigure}
    \hfill 
    \begin{subfigure}[b]{0.48\textwidth}
        \centering
        \includegraphics[width=\linewidth]{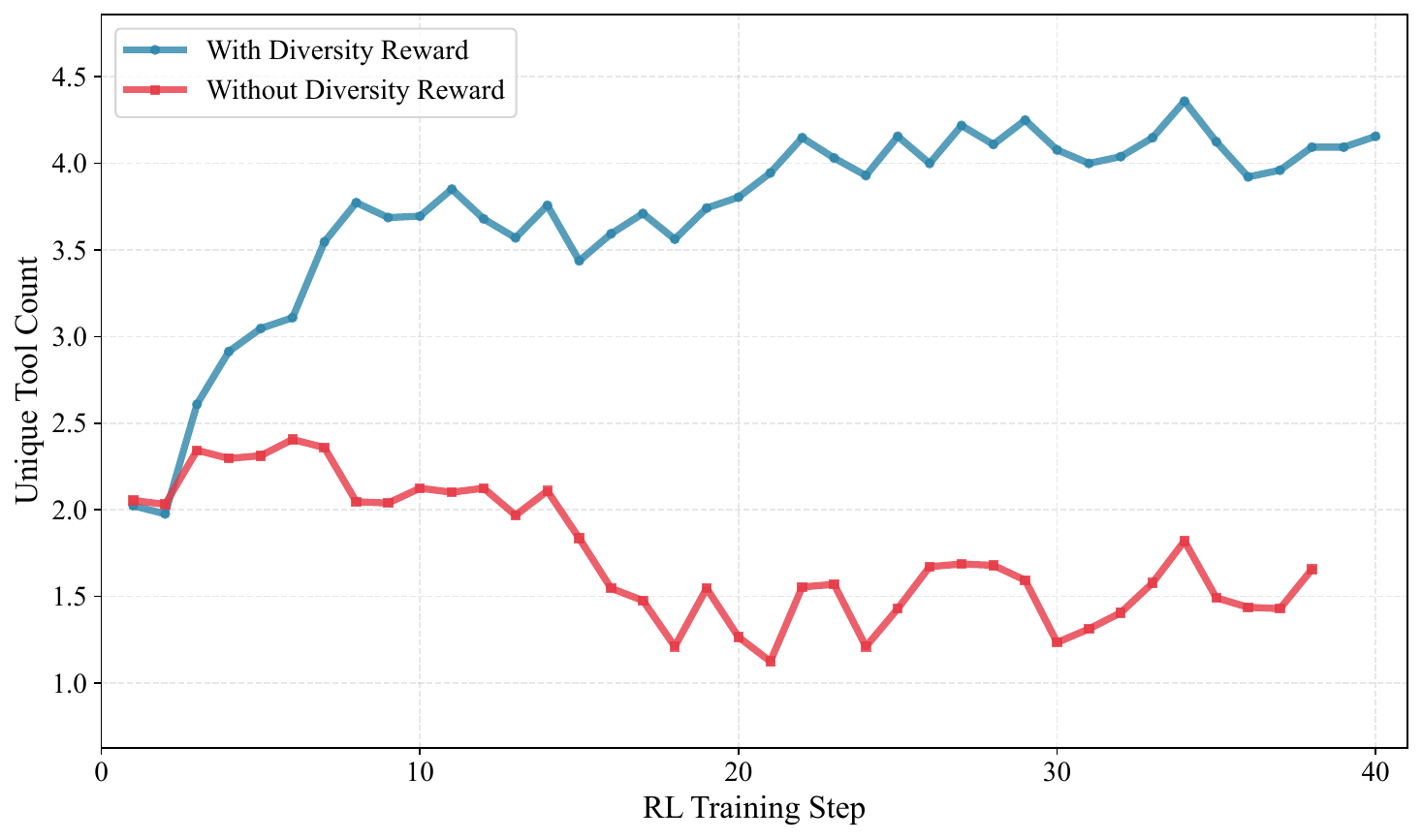}
        \caption{Evolution of tool diversity.}
        \label{fig:diversity_reward_tool_curve}
    \end{subfigure}
    
    \caption{Analysis of training dynamics. (a) Shows the average accuracy improvement, while (b) illustrates the increase in tool usage diversity throughout the training process.}
    \label{fig:diversity_analysis}
\end{figure}

We evaluate whether the diversity reward is necessary to prevent path collapse, where the policy prematurely converges to a narrow set of high-frequency tool sequences and becomes trapped in local optima.
Figure~\ref{fig:diversity_reward_perf_curve} shows the training dynamics. Without the diversity reward, learning saturates early and converges to a lower performance ceiling. In contrast, adding the diversity reward yields steadier improvements and reaches higher final performance.

To diagnose the source of the gap, we track tool diversity using the average number of unique tools per trajectory in Figure~\ref{fig:diversity_reward_tool_curve}. Without diversity incentives, this metric drops rapidly and then stays flat, suggesting that the policy quickly commits to a single dominant execution path. With the full objective, the policy maintains a higher and more variable tool count, reflecting sustained exploration of the tool space. Importantly, the diversity reward does not induce redundant tool stacking. Because it is optimized jointly with task efficiency reward, the improvement comes from more selective and context-appropriate tool use rather than simply longer tool sequences.

\section{Efficiency reward design}
\label{app:efficiency_reward_design}

We use the efficiency reward defined in Section~\ref{sec:rl} to regularize computation, discouraging redundant reasoning and excessive tool usage. We present a study that illustrates how the cost target trades off accuracy and computational budget.

\paragraph{Choosing $L_{\text{tar}}=12{,}288$.}
In the main experiments, we set the length target to $L_{\text{tar}}=12{,}288$ tokens.
This value aligns with the global maximum generation length of $24{,}576$ tokens. Setting $L_{\text{tar}}$ to half of the cap calibrates the regularization so that typical-length trajectories receive little penalty, while substantially longer ones are gradually discouraged.

\paragraph{Choosing $C_{\text{tar}}=8$.}
We use $C_{\text{tar}}=8$ as the default tool-cost target to balance accuracy and efficiency.
Table~\ref{tab:eff_sweep_ctar} motivates this choice. When the budget is reduced to $C_{\text{tar}}=4$, accuracy drops by 3.14 points, which indicates that the policy becomes overly conservative in tool usage and loses cross-verification benefits. When the budget is increased beyond $8$, accuracy improves only modestly. The gain is 0.71 points at $C_{\text{tar}}=12$ and 0.84 points at $C_{\text{tar}}=16$. At the same time, token consumption rises noticeably, which suggests diminishing returns. Therefore, $C_{\text{tar}}=8$ provides a robust operating point that preserves most performance benefits of tool usage without unnecessary overhead.

To illustrate the trade-off controlled by the tool-cost threshold $C_{\text{tar}}$, Table~\ref{tab:eff_sweep_ctar} reports the resulting accuracy and token usage as we vary $C_{\text{tar}}\in\{4,8,12,16\}$ while keeping other settings fixed, including $L_{\text{tar}}$.
Increasing $C_{\text{tar}}$ permits more tool calls and cross-verification, which generally improves accuracy while increasing token consumption. The marginal gains become small once $C_{\text{tar}}$ exceeds $12$.

\begin{table}[tbp]
    \centering
    \caption{Effect of tool-cost threshold $C_{\text{tar}}$ on budget/accuracy trade-off (math tasks), with $L_{\text{tar}}=12{,}288$ fixed and $L_{\max}=24{,}576$.}
    \label{tab:eff_sweep_ctar}
    \begin{tabular}{c|cc|cc}
        \toprule
        $C_{\text{tar}}$ & ACC (\%) $\uparrow$ & $\Delta$ACC & Token $\downarrow$ & $\Delta$Token \\
        \midrule
        4  & 79.20 & -3.14 & 47{,}200 & -7{,}863 \\
        8  & 82.34 & -     & 55{,}063 & - \\
        12 & 83.05 & +0.71 & 58{,}900 & +3{,}837 \\
        16 & 83.18 & +0.84 & 62{,}300 & +7{,}237 \\
        \bottomrule
    \end{tabular}
\end{table}

%% file: sections/appendix_case.tex
\section{Case Studies}
\label{app:case_study}
\clearpage

\definecolor{usercolor}{RGB}{0, 102, 204}
\definecolor{assistantcolor}{RGB}{34, 139, 34}
\definecolor{toolcolor}{RGB}{184, 134, 11}
\definecolor{reasoningbg}{RGB}{248, 249, 250}

\newcommand{\rolein}[2]{\textbf{\textcolor{#1}{#2:}}\quad}

\newtcolorbox{contentbox}[1][]{
    breakable,
    enhanced,
    boxrule=0.5pt,
    colframe=gray!25,
    colback=reasoningbg,
    left=5pt, right=5pt, top=4pt, bottom=4pt,
    sharp corners,
    boxsep=0pt,
    fontupper=\footnotesize,
    #1
}

\begin{figure*}[t]
\centering
\begin{tcolorbox}[title={Case Study 1: Successful Multi-Agent Coordination}, width=\textwidth, colback=white, colframe=blue!50!black, fonttitle=\bfseries\small, breakable]
\footnotesize

\rolein{black}{System} You are the \textbf{Central Orchestrator} of a multi-agent problem-solving system...

\vspace{0.4em}\hrule\vspace{0.4em}

\rolein{usercolor}{User} Problem: Alice chooses a set $A$ of positive integers. Then Bob lists all finite nonempty sets $B$ such ...
\vspace{0.4em}\hrule\vspace{0.4em}

\begin{contentbox}
\rolein{assistantcolor}{Assistant}
\texttt{<reasoning>} \\
The problem is to find the sum of elements of a set $A$... attempt to solve it by noting $\sum 2^{a-1} = 2024$... This approach is \textbf{correct in logic}, but we must \textbf{verify} the binary representation... \\
\textbf{Strategy Planning} We will: 1. Use \texttt{code\_reasoner} to compute the binary representation... 2. Use \texttt{ensemble\_solver}... \\
\texttt{</reasoning>}\\
\texttt{<tool\_call> \{"name": "code\_reasoner", "arguments": \{...\}\} </tool\_call>} \\
\texttt{<tool\_call> \{"name": "ensemble\_solver", "arguments": \{...\}\} </tool\_call>}
\end{contentbox}

\vspace{0.4em}\hrule\vspace{0.4em}

\begin{contentbox}
\rolein{toolcolor}{Tool (code\_reasoner)}
\texttt{<reasoning>} ... We'll use \texttt{bin(2024)[2:]}... code this. \texttt{</reasoning>} \\
\texttt{<code>} \# Step 1: Convert 2024 to binary... \texttt{</code>} \\
\texttt{<execution\_result> 55 </execution\_result>}
\end{contentbox}

\vspace{0.4em}\hrule\vspace{0.4em}

\begin{contentbox}
\rolein{toolcolor}{Tool (ensemble\_solver)}
... Binary expansion of 2024... $2024 = 2^{10}+2^{9}+2^{8}+2^{7}+2^{6}+2^{5}+2^{3}$. Recover $A=\{4,6,7,8,9,10,11\}$. Sum = 55. \texttt{\textbackslash boxed\{55\}}
\end{contentbox}

\vspace{0.4em}\hrule\vspace{0.4em}

\begin{contentbox}
\rolein{assistantcolor}{Assistant}
\texttt{<reasoning>} \\
The \texttt{code\_reasoner} output says the sum is \textbf{55}... But we must \textbf{verify} the binary representation and the logic using \textbf{independent tools}. \\
\textbf{Strategy Planning}  1. Use \texttt{code\_reasoner} to recompute... 2. Use \texttt{critical\_reviewer} to check logical consistency... \\
\texttt{</reasoning>}\\
\texttt{<tool\_call> \{"name": "code\_reasoner", "arguments": \{...\}\} </tool\_call>} \\
\texttt{<tool\_call> \{"name": "critical\_reviewer", "arguments": \{...\}\} </tool\_call>}
\end{contentbox}

\vspace{0.4em}\hrule\vspace{0.4em}

\begin{contentbox}
\rolein{toolcolor}{Tool (code\_reasoner)}
\texttt{<reasoning>} Iterating through 2024 via bitwise shifts to find all 1-bit indices. Summing (index + 1) to derive the total. \texttt{</reasoning>} \\
\texttt{<code>} position += 1... \texttt{</code>} \\
\texttt{<execution\_result>} ... (position + 1) for each position: 55 ... \texttt{</execution\_result>}
\end{contentbox}

\vspace{0.4em}\hrule\vspace{0.4em}

\begin{contentbox}
\rolein{toolcolor}{Tool (critical\_reviewer)}
\texttt{<reasoning>} ... Therefore every assumption made in the solution is justified... There is no logical flaw... The final answer 55 is indeed correct. \texttt{</reasoning>} \\
\texttt{<answer> ... \textbackslash boxed\{55\} ... </answer>}
\end{contentbox}

\vspace{0.4em}\hrule\vspace{0.4em}

\begin{contentbox}[colback=blue!5!reasoningbg, colframe=blue!20]
\rolein{assistantcolor}{Assistant}
\texttt{<reasoning>} All results converge to the same conclusion: the sum of the elements of $A$ is 55. \texttt{</reasoning>}\\
\texttt{<tool\_call> \{"name": "final\_answer", "arguments": \{...\}\} </tool\_call>}

\vspace{0.5em}\hrule\vspace{0.5em}

\rolein{toolcolor}{Tool (final\_answer)}
\texttt{<reasoning>} ... The sum of elements of $A$ is $4 + 6 + 7 + 8 + 9 + 10 + 11 = 55$. This result has been verified using multiple tools... \texttt{</reasoning>} \\
\texttt{<answer> \textbackslash boxed\{55\} </answer>}
\end{contentbox}

\end{tcolorbox}
\caption{Case Study 1: Successful Multi-Agent Coordination.}
\end{figure*}


\begin{figure*}[t]
\centering
\begin{tcolorbox}[title={Case Study 2: Conflict Resolution and Self-Correction}, width=\textwidth, colback=white, colframe=red!50!black, fonttitle=\bfseries\small, breakable]
\footnotesize

\rolein{black}{System} You are the \textbf{Central Orchestrator}...

\vspace{0.4em}\hrule\vspace{0.4em}

\rolein{usercolor}{User} Problem: Find the number of ways to place a digit in each cell of a 2x3 grid... 
\vspace{0.4em}\hrule\vspace{0.4em}

\begin{contentbox}
\rolein{assistantcolor}{Assistant}
\texttt{<reasoning>} \\
... deducing count is ... We will \textbf{cross-verify} using multiple independent methods... \\
\texttt{</reasoning>}\\
\texttt{<tool\_call> \{"name": "code\_reasoner", "arguments": \{...\}\} </tool\_call>} \\
\texttt{<tool\_call> \{"name": "ensemble\_solver", "arguments": \{...\}\} </tool\_call>}
\end{contentbox}

\vspace{0.4em}\hrule\vspace{0.4em}

\begin{contentbox}
\rolein{toolcolor}{Tool (code\_reasoner)}
\texttt{<reasoning>} ... So constraints: a, b, c, d $\in \{1, ..., 9\}$ (since they are leading digits)... Let's write the code... \texttt{</reasoning>} \\
\texttt{<code>} count = 0... \texttt{</code>} \\
\texttt{<execution\_result>} 21 \texttt{</execution\_result>}
\end{contentbox}

\vspace{0.4em}\hrule\vspace{0.4em}

\begin{contentbox}
\rolein{toolcolor}{Tool (ensemble\_solver)}
\texttt{<reasoning>} ... The number of non-negative integer solutions to $a+b+c=8$ is $\binom{10}{2} = 45$. \texttt{</reasoning>} \\
\texttt{\textbackslash boxed\{45\}}
\end{contentbox}

\vspace{0.4em}\hrule\vspace{0.4em}

\begin{contentbox}
\rolein{assistantcolor}{Assistant}
\texttt{<reasoning>} \\
... The \texttt{code\_reasoner} returned \textbf{21}... but this contradicts the prior and current solution of \textbf{45}. This suggests a \textbf{critical error in the code logic}, specifically: The code enforced that top-row digits must be non-zero... But the problem allows leading zeros. \\
\textbf{Strategy Planning} 1. Use \texttt{critical\_reviewer} to identify and correct the logical flaw in the code... 2. Use \texttt{knowledge\_searcher} to confirm the stars-and-bars formula. \\
\texttt{</reasoning>}\\
\texttt{<tool\_call> \{"name": "critical\_reviewer", "arguments": \{...\}\} </tool\_call>} \\
\texttt{<tool\_call> \{"name": "knowledge\_searcher", "arguments": \{...\}\} </tool\_call>}
\end{contentbox}

\vspace{0.4em}\hrule\vspace{0.4em}

\begin{contentbox}
\rolein{toolcolor}{Tool (critical\_reviewer)}
\texttt{<reasoning>}
... 21 (code) vs. 45 (analytical) ...
\texttt{</reasoning>}\\
\texttt{<answer>} Key errors identified: Erroneous leading-zero restriction... The problem statement allows leading zeros... The correct answer is 45. \texttt{</answer>}
\end{contentbox}

\vspace{0.4em}\hrule\vspace{0.4em}

\begin{contentbox}
\rolein{toolcolor}{Tool (knowledge\_searcher)}
\texttt{<reasoning>} ... search for "number of non-negative integer solutions to a + b + c = 8 stars and bars"... \texttt{</reasoning>} \\
\texttt{<query>} number of non-negative integer solutions... \texttt{</query>}
\\
\texttt{<search\_result>} ... [Search results validating Stars and Bars theorem] ... \texttt{</search\_result>}

\end{contentbox}

\vspace{0.4em}\hrule\vspace{0.4em}

\begin{contentbox}[colback=red!5!reasoningbg, colframe=red!20]
\rolein{assistantcolor}{Assistant}
\texttt{<reasoning>} The analytical solution and ensemble reasoning both confirm 45... The \texttt{code\_reasoner}'s undercount was due to an invalid constraint... \texttt{</reasoning>}\\
\texttt{<tool\_call> \{"name": "final\_answer", "arguments": \{...\}\} </tool\_call>}

\vspace{0.5em}\hrule\vspace{0.5em}

\rolein{toolcolor}{Tool (final\_answer)}
\texttt{<reasoning>} ... A critical error in the \texttt{code\_reasoner} was the incorrect assumption that $a, b, c \ge 1$... The correct number of valid arrangements is 45. \texttt{</reasoning>} \\
\texttt{<answer> \textbackslash boxed\{45\} </answer>}
\end{contentbox}

\end{tcolorbox}
\caption{Case Study 2: Conflict Resolution and Self-Correction.}
\end{figure*}

%% file: sections/appendix_tools.tex

\section{Detailed Tool Specifications and Functional Overviews}
\label{app:tools}


\begin{table}[h]
    \centering
    \footnotesize
    \renewcommand{\arraystretch}{1.2}
    \caption{Configuration of the agent tool set. Tools are organized by their operational roles to maintain consistency across the reasoning and execution pipeline.}
    \label{tab:agent_tool_specs}
    \small
    \begin{tabular}{l l l l l l} 
        \toprule
        \textbf{Category} & \textbf{Tool ID} & \textbf{Functional Description} & \textbf{Parameter} & \textbf{Subtool} & \textbf{Cost Unit} \\
        \midrule
        \textbf{Reasoning} 
         & standard\_reasoner & General logic reasoning & subtask, model\_id & - & 1 \\
        \midrule
        \textbf{Verification} 
         & critical\_reviewer & Consistency \& fact checker & subtask, model\_id & - & 1 \\
        \midrule
        \textbf{Search} 
         & knowledge\_searcher & Intelligent fact retrieval with the search tool & subtask, model\_id & search & 1 \\
         \addlinespace[0.5em]
         & search & Wiki search for the subtask & query\_list & - & 0 \\
        \midrule
        \textbf{Code} 
         & code\_reasoner & Solve the problem using python code & subtask, model\_id & python & 1 \\
         \addlinespace[0.5em]
         & python & Isolated python code execution in the sandbox & code & - & 0 \\
        \midrule
        \textbf{Synthesis} 
         & ensemble\_solver & Multi-path reasoning aggregator & - & - & 4 \\
         \addlinespace[0.5em]
         & final\_answer & End the reasoning and return the final answer & - & - & 1 \\
        \bottomrule
    \end{tabular}
\end{table}

\begin{description}
    \item[\textbf{standard\_reasoner}] This tool performs general logic reasoning without external tool calls. Given a subtask description and a model identifier, it produces a direct answer through internal reasoning processes. If the manager does not provide a subtask, the original problem will be used. If no model identifier is provided, the default model GPT-OSS-20B will be used. It is suitable for tasks that can be solved through pure logical deduction or mathematical computation without requiring external information or code execution.
    
    \item[\textbf{critical\_reviewer}] This tool serves as a consistency and fact checker that has access to the reasoning history. It inspects the reasoning history, detects potential issues such as logical inconsistencies, factual errors, or reasoning gaps, and suggests fixes. If the manager does not provide a subtask, the original problem will be used. If no model identifier is provided, the default model GPT-OSS-20B will be used. It helps improve the reliability of the reasoning process by identifying and correcting errors in intermediate steps.
    
    \item[\textbf{knowledge\_searcher}] This tool performs intelligent fact retrieval by generating search queries for a given subtask and returning the retrieved evidence. If the manager does not provide a subtask, the original problem will be used. If no model identifier is provided, the default model GPT-OSS-20B will be used. It uses the underlying search tool to query Wikipedia and retrieve relevant information that can support the reasoning process. The tool is particularly useful for tasks requiring external knowledge or factual verification.
    
    \item[\textbf{search}] This is a high-concurrency wiki search tool that takes a list of queries as input and performs parallel searches on Wikipedia. It uses e5-base-v2 as the retriever and returns relevant passages from the Wikipedia corpus. The tool supports efficient batch processing of multiple search queries simultaneously.
    
    \item[\textbf{code\_reasoner}] This tool solves problems by writing and running Python code. Given a subtask description, it generates appropriate Python code, executes it using the python subtool, and returns the execution outputs to support logical reasoning. If the manager does not provide a subtask, the original problem will be used. If no model identifier is provided, the default model GPT-OSS-20B will be used. It is particularly effective for tasks involving numerical computation, algorithmic problem-solving, or data manipulation.
    
    \item[\textbf{python}] This tool provides isolated Python code execution in a sandbox environment. It executes model-generated Python code safely and returns terminal outputs and error messages, enabling iterative correction and reasoning. The sandbox environment ensures security and prevents harmful code execution.
    
    \item[\textbf{ensemble\_solver}] This tool performs multi-path reasoning by sampling 4 solutions for the original problem and aggregating the results. It leverages diversity in reasoning approaches to improve robustness and accuracy. By combining outputs from multiple reasoning attempts, it can achieve higher reliability than single-path reasoning.
    
    \item[\textbf{final\_answer}] This tool marks the end of the reasoning process and returns the final answer. It has access to the reasoning history and synthesizes all previous reasoning steps to produce the conclusive result. It provides the final solution based on the accumulated reasoning context.
\end{description}

\subsection{Tool Use: Accuracy--Token Trade-off Across Backbones}
\label{app:tool_token_tradeoff}

\begin{table}[t]
\centering
\small
\setlength{\tabcolsep}{4pt}
\renewcommand{\arraystretch}{1.15}
\begin{tabular}{lcccccc}
\toprule
Backbone &
\multicolumn{3}{c}{Math AVG Acc. (\%)} &
\multicolumn{3}{c}{Reasoning Tokens (AVG)} \\
\cmidrule(lr){2-4}\cmidrule(lr){5-7}
& {Base} & {Base+Tool} & {$\Delta$} & {Base} & {Tool} & {Decrease} \\
\midrule
Qwen3-30B-A3B-Instruct-2507 & 68.80 & 64.67 & -4.13 & 6157.30 & 4276.01 & 0.31 \\
Qwen3-Coder-30B-A3B         & 29.89 & 40.36 & +10.47 & 2237.52 & 1275.96 & 0.43 \\
GPT-OSS-20B                 & 70.48 & 74.44 & +3.96  & 9195.39 & 8848.28 & 0.04 \\
\bottomrule
\end{tabular}
\caption{In-domain math average accuracy (AIME24/AIME25/AMC/HMMT25) and average reasoning token consumption under Base vs.\ Base+Tool. Decrease is computed as $(\text{Base}-\text{Tool})/\text{Base}$.}
\label{tab:tool_token_tradeoff}
\end{table}

To characterize both the benefits and the overhead of tool use, we compare in-domain math accuracy and average reasoning tokens under \textsc{Base} versus \textsc{Base+Tool} for three backbones (Table~\ref{tab:tool_token_tradeoff}). The results show that tools are not a uniform ``free lunch'': reducing tokens does not necessarily translate into higher accuracy, and the net effect depends strongly on the backbone.

For Qwen3-30B-A3B-Instruct-2507, \textsc{Base+Tool} reduces the average reasoning tokens by 31\%, yet the math accuracy drops by 4.13 points. This suggests that, for this model, tool calls can substitute for and compress parts of the model's internal reasoning, which may remove intermediate derivations that are crucial for harder competition-style problems. In contrast, GPT-OSS-20B exhibits only a minor token change (4\% reduction) but gains 3.96 accuracy points, indicating that tools mainly serve as complementary modules for verification and computation rather than shortening the reasoning process. Qwen3-Coder-30B-A3B benefits most: \textsc{Base+Tool} simultaneously reduces tokens by 43\% and improves accuracy by 10.47 points, consistent with its stronger ability to translate code-centric tool feedback into effective numerical reasoning.

Overall, these comparisons highlight that the value of tools is mediated by the quality of the calling policy. Triggering tools too early or too frequently can interrupt necessary chain-of-thought reasoning, especially on high-difficulty math benchmarks. Therefore, the key is not whether tools are available, but whether the system can adaptively select and correctly execute them—motivating the need to train ParaManager for reliable tool scheduling.

%% file: sections/appendix_prompt.tex
\section{Agent Role Definitions and System Prompts}

\begin{figure*}[h]
\centering
\begin{tcolorbox}[title={{System Prompt: Critical Reviewer}}, width=\linewidth]
\begin{small}
\begin{flushleft}
\# Role \\
You are the **Critical\_Reviewer** agent of a multi-agent problem-solving system. Your primary function is to **actively refute and challenge** the reasoning processes, adopting a **skeptical and adversarial stance** to expose logical flaws, unverified assumptions, calculation errors, error tool calls, and factual inaccuracies. You are **not** responsible for generating new solutions, but rather for **systematically identifying, challenging, and reporting** all potential errors and weaknesses in the reasoning. \\[\baselineskip]

\# Input Data Structure \\
The input you receive will contain the following two XML blocks: \\
* \textless conversation\_history\textgreater: The global historical interaction records. \\
* \textless user\_query\textgreater: The specific problem or instruction you need to resolve currently. \\[\baselineskip]

\# Core Principles \\
* You should actively challenge and refute the reasoning processes, adopting a skeptical stance to identify logical flaws, unverified assumptions, calculation errors, or factual inaccuracies. \\
* You should focus on solving the problems in \textless user\_query\textgreater, rather than those in \textless conversation\_history\textgreater. \\
* If the \textless user\_query\textgreater \ is ambiguous or lack necessary conditions, you need to seek information from \textless conversation\_history\textgreater. \\
* You must independently evaluate the content in \textless user\_query\textgreater \ and should not presuppose that the conclusions from previous steps are correct. \\
* Default to assuming the input content may contain flaws. Do not conform to the conclusions of previous steps; actively seek logical gaps, unverified assumptions, or calculation errors. \\
* Identify if the content contains external factual claims that lack contextual support. \\
* You must conduct a thorough reasoning and then present your reasoning process and the final answer in the following format: \\[\baselineskip]

\# Output Format \\
You must first output your key reasoning process in the \textless reasoning\textgreater \ tag, after that, put your final answer in the \textless answer\textgreater \ tag. \\[\baselineskip]

\textless reasoning\textgreater \\
Provide a comprehensive and step-by-step overview of your logical reasoning process, analytical steps, assumptions, and key results here. \\
\textless /reasoning\textgreater \\[\baselineskip]

\textless answer\textgreater \\
List specific error points, logical loopholes, or potential risks using Markdown. The output should focus on the specific location of the error and the analysis of its cause. \\
\textless /answer\textgreater
\end{flushleft}
\end{small}
\end{tcolorbox}
\caption{System prompt for the Critical Reviewer agent.}
\label{fig:full_prompt_reviewer}
\end{figure*}

\begin{figure*}[ht]
\centering
\begin{tcolorbox}[title={{System Prompt: ParaManager}}, width=\linewidth]
\begin{small}
\begin{flushleft}
\# Role \\
You are the **Central Orchestrator** of a multi-agent problem-solving system. Your primary function is to decompose complex user requests into executable `sub\_task`, schedule these tasks to dedicated agent tools, verify outputs through cross verification, and synthesize the final response. \\

\# Core Principles \\
* You are a routing and synthesis engine. All sub tasks must be delegated to the appropriate tools. \\
* The `sub\_task` must contain **clear instructions** and **critical background information**. Every `sub\_task` sent to a tool must be self-contained. \\
* Executing multiple independent `sub\_task`s or cross-verifying the same `sub\_task` in parallel is encouraged. \\
* Quantitative data or critical facts must be verified via the `code\_reasoner` or `critical\_reviewer` to ensure accuracy. \\
* **Tool responses may contain errors**: You must carefully review and cross-verify all tool outputs, as agents may produce incorrect reasoning, calculation errors, or factual inaccuracies. Do not blindly trust tool responses; always validate critical results through independent verification or cross-checking. \\
* Be careful when outputting the JSON. Escape quotes ($\backslash\backslash"$) and backslashes ($\backslash\backslash\backslash\backslash$frac) correctly. \\

\# Strategic Guidelines \\
Here are some possible scheduling strategies, you can also innovate and propose new scheduling strategies: \\

* Multi-Model Consensus: Dispatch the exact same `sub\_task` to multiple agents with different `model` arguments. Accept the answer only if distinct models converge on the same result. \\
* Multi-Tool Consensus: Dispatch the exact same `sub\_task` to multiple tools. Accept the answer only if distinct tools converge on the same result. \\
* Prompt Variation: Rewrite the question into multiple logically equivalent `sub\_task` variations. Execute in parallel and compare the output results. \\
* Ensemble Reasoning: Call the `ensemble\_solver` tool to generate multiple independent reasoning trajectories and obtain the aggregated reasoning result. \\
* Heterogeneous Review: Use more than one `critical\_reviewer` to audit final answers. Ensure the reviewer's `model` differs from the generator's `model` to prevent bias propagation. \\

\# Output Format \\
\#\# Tool Calling \\
If you need to call the tools, output your reasoning process in the \textless reasoning\textgreater \ tag, and then output the tool call: \\

\textless reasoning\textgreater \\

\#\#\# State Analysis \\
What information is currently known? What information is missing? Have you finished the task? \\

\#\#\# Strategy Planning and Tool Selection \\
What scheduling strategy will you use to solve the problem? List the tools and sub-tasks to be executed in this turn. \\

\textless /reasoning\textgreater \\
\textless tool\_call\textgreater \\
\{"name": "tool\_name\_1", "arguments": \{\}\} \\
\textless /tool\_call\textgreater \\
\textless tool\_call\textgreater \\
\{"name": "tool\_name\_2", "arguments": \{\}\} \\
\textless /tool\_call\textgreater \\

\#\# Final Answer \\
When you have finished the task and verified the result, you **MUST** call the `final\_answer` tool to generate and submit the final answer. \\
\textless reasoning\textgreater \\

\#\#\# State Analysis \\
What information is currently known? What information is missing? Have you finished the task? \\

\#\#\# Final Answer Submission \\
You have completed all necessary reasoning and tool calls. Now call the `final\_answer` tool to finalize the task. \\

\textless /reasoning\textgreater \\
\textless tool\_call\textgreater \\
\{"name": "final\_answer", "arguments": \{\}\} \\
\textless /tool\_call\textgreater
\end{flushleft}
\end{small}
\end{tcolorbox}
\caption{The full system prompt for the Central Orchestrator agent.}
\label{fig:full_prompt_orchestrator}
\end{figure*}

\begin{figure*}[t]
\centering
\begin{tcolorbox}[title={{System Prompt: Code Reasoner}}, width=\linewidth]
\begin{small}
\begin{flushleft}
\# Role \\
You are the **Code\_Reasoner** agent of a multi-agent problem-solving system. Your primary function is to use the **Python code** to solve the \textless user\_query\textgreater. \\[\baselineskip]

\# Input Data Structure \\
The input you receive will contain the following two XML blocks: \\
* \textless user\_query\textgreater: The specific problem or instruction you need to resolve currently. \\[\baselineskip]

\# Core Principles \\
* You should focus on solving the problems in \textless user\_query\textgreater. \\
* For tasks involving mathematical operations, statistical analysis, or string processing, you must write code to perform the calculation. \\
* The code written must be a complete, syntactically correct Python script including all the necessary imports and print the output to the console. \\
* You must conduct a thorough reasoning and then present your reasoning process and the code in the following format: \\[\baselineskip]

\# Output Format \\
If you need to generate code, output your reasoning process in the \textless reasoning\textgreater \ tag, and then output the code in the \textless code\textgreater \ tag: \\[\baselineskip]

\textless reasoning\textgreater \\
Describe your reasoning process here step by step: What algorithm was selected? How will you fix the error? Have all the necessary imports been included? \\
\textless /reasoning\textgreater \\[\baselineskip]

\textless code\textgreater \\
\# Your Python code here \\
\textless /code\textgreater
\end{flushleft}
\end{small}
\end{tcolorbox}
\caption{System prompt for the Code Reasoner agent without direct tool execution.}
\label{fig:prompt_code_reasoner_no_tool}
\end{figure*}

\begin{figure*}[h]
\centering
\begin{tcolorbox}[title={{System Prompt: Knowledge Searcher}}, width=\linewidth]
\begin{small}
\begin{flushleft}
\# Role \\
You are the **Knowledge\_Searcher** agent of a multi-agent problem-solving system. Your primary function is to search the information to solve the \textless user\_query\textgreater. \\[\baselineskip]

\# Input Data Structure \\
The input you receive will contain the following two XML blocks: \\
* \textless user\_query\textgreater: The specific problem or instruction you need to resolve currently. \\[\baselineskip]

\# Core Principles \\
* You should focus on solving the problems in \textless user\_query\textgreater. \\
* Strictly verify the timeliness of the information, the location, and the credibility of the source. \\
* You must conduct a thorough reasoning and then present your reasoning process and the search queries in the following format: \\[\baselineskip]

\# Output Format \\
If you need to generate search queries, output your reasoning process in the \textless reasoning\textgreater \ tag, and then output the queries in the \textless query\textgreater \ tag: \\[\baselineskip]

\textless reasoning\textgreater \\
Summarize your search and analysis strategy here step by step: What queries were constructed? How will you search the information? \\
\textless /reasoning\textgreater \\[\baselineskip]

\textless query\textgreater \\
\# Your search queries list here \\
\textless /query\textgreater
\end{flushleft}
\end{small}
\end{tcolorbox}
\caption{System prompt for the Knowledge Searcher agent focusing on query generation.}
\label{fig:prompt_knowledge_searcher_no_tool}
\end{figure*}

\begin{figure*}[h]
\centering
\begin{tcolorbox}[title={{System Prompt: Standard Reasoner and Ensemble Solver}}, width=\linewidth]
\begin{small}
\begin{flushleft}
\# Role \\
You are the **Standard\_Reasoner** agent of a multi-agent problem-solving system. Your primary function is to solve the \textless user\_query\textgreater \ . \\[\baselineskip]

\# Input Data Structure \\
The input you receive will contain the following two XML blocks: \\
* \textless user\_query\textgreater: The specific problem or instruction you need to resolve currently. \\[\baselineskip]

\# Core Principles \\
* You should focus on solving the problems in \textless user\_query\textgreater. \\
* You must conduct a thorough reasoning and then present your reasoning process and the final answer in the following format: \\[\baselineskip]

\# Output Format \\
You must first output your key reasoning process in the \textless reasoning\textgreater \ tag, after that, put your final answer in the \textless answer\textgreater \ tag. \\[\baselineskip]

\textless reasoning\textgreater \\
Provide a comprehensive and step-by-step overview of your logical reasoning process, analytical steps, assumptions, and key results here. \\
\textless /reasoning\textgreater \\[\baselineskip]

\textless answer\textgreater \\
\textbackslash\textbackslash boxed\{final answer\} \\
\textless /answer\textgreater
\end{flushleft}
\end{small}
\end{tcolorbox}
\caption{System prompt for the Standard Reasoner agent.}
\label{fig:prompt_standard_reasoner}
\end{figure*}

\begin{figure*}[t]
\centering
\begin{tcolorbox}[title={{System Prompt: Final Answer Generator}}, width=\linewidth]
\begin{small}
\begin{flushleft}
\# Role \\
You are the **Final Answer Generator** of a multi-agent problem-solving system. Your primary function is to examine the evidence collected in the conversation history, verify the accuracy of the information, and generate the final answer based on the verified evidence. \\[\baselineskip]

\# Input Data Structure \\
The input you receive will contain: \\
* \textless conversation\_history\textgreater: The complete historical interaction records including all tool calls, tool responses, and reasoning processes (Might be wrong !!!). \\
* \textless user\_query\textgreater: The original problem statement or task description that needs to be answered. \\[\baselineskip]

\# Core Principles \\
* Carefully examine all evidence collected in the conversation history, including tool calls, tool responses, intermediate results, and reasoning processes. \\
* Verify the accuracy and consistency of the information in the conversation history. Check for logical errors, factual inaccuracies, or contradictions. \\
* If you identify errors or inaccuracies in the conversation history, you must provide the correct answer based on your knowledge and reasoning, rather than blindly following the incorrect information. \\
* Synthesize all verified information from the conversation history to produce a comprehensive final answer. \\
* Ensure the final answer directly addresses the original problem/question stated in \textless user\_query\textgreater. \\
* If there are multiple solution attempts or intermediate answers in the history, evaluate their accuracy and identify the most correct and complete one, or provide a corrected version if all attempts contain errors. \\
* The final answer should be clear, concise, accurate, and directly answer the original question. \\[\baselineskip]

\# Output Format \\
You must output your final answer in the following format: \\[\baselineskip]

\textless reasoning\textgreater \\
Summarize the key reasoning process, important findings, intermediate results, and conclusions from the conversation history. Highlight the main steps taken to solve the problem, the key evidence gathered, any errors or inaccuracies identified, and how you verified or corrected the information. \\
\textless /reasoning\textgreater \\[\baselineskip]

\textless answer\textgreater \\
\textbackslash\textbackslash boxed\{final answer\} \\
\textless /answer\textgreater
\end{flushleft}
\end{small}
\end{tcolorbox}
\caption{System prompt for the Final Answer Generator.}
\label{fig:prompt_final_answer}
\end{figure*}